\pdfoutput=1

\documentclass[11pt]{article}

\usepackage[final]{acl}

\usepackage{times}
\usepackage{latexsym}
\usepackage[many]{tcolorbox}
\usepackage[T1]{fontenc}
\usepackage[utf8]{inputenc}
\usepackage{microtype}
\usepackage{inconsolata}
\usepackage{graphicx}
\usepackage{microtype}

\usepackage{amsmath}
\usepackage{amssymb}
\usepackage{mathtools}
\usepackage{amsthm}
\usepackage{multirow}
\usepackage{makecell}
\usepackage{booktabs}
\usepackage{array}
\usepackage{longtable}
\usepackage{subcaption}
\usepackage{fontawesome}
\usepackage{centernot}

\theoremstyle{plain}
\newtheorem{theorem}{Theorem}[section]

\theoremstyle{definition}
\newtheorem{definition}[theorem]{Definition}

\theoremstyle{remark}

\newenvironment{findingBox}[2]{%
	\begin{tcolorbox}[
colframe=black!80,
colback=gray!10,
 boxrule=.5pt,
 left=1pt,
 right = 1pt,
 top=0pt,
 bottom=0pt,
 size=small,
 fonttitle=\bfseries,
coltitle=black,
boxrule=0.4mm,
arc=2mm
 ]{\textbf{Finding #1:} #2} 
}{%
	\end{tcolorbox}
}

\newtcolorbox{prompt}[2][]{
    colback=white,
    colframe=gray!45,
    fonttitle=\bfseries,
    coltitle=black,
    sharp corners,
    title=#2,
    #1
}

\tcbset{
    promptstyle/.style={
        enhanced,
        colback=white,
        colframe=black,
        colbacktitle=gray!20,
        width=0.98\linewidth, 
        coltitle=black,
        rounded corners,
        sharp corners=north,
        boxrule=0.5pt,
        drop shadow=black!50!white,
        attach boxed title to top left={
            xshift=-2mm,
            yshift=-2mm
        },
        title code={%
            \begin{minipage}{0.8\linewidth}
                \centering\thetitle
            \end{minipage}
        },
        boxed title style={
            rounded corners,
            size=small,
            colback=gray!20,
        },
        fonttitle=\normalsize\bfseries,
        before upper={\parindent15pt}
    }
}

\newtcolorbox{promptbox}[1][]{
    promptstyle,
    title=Prompt,
    #1
}

\title{
\vspace{-2em}{\small \hfill NAACL 2025 Main Conference}\\
\vspace*{.5in}
Stronger Models are Not Always Stronger Teachers for Instruction Tuning}

\vspace{-0.5em}
\author{
\textbf{Zhangchen Xu}\;\;\; 
\textbf{Fengqing Jiang} \;\;\; 
\textbf{Luyao Niu}\;\;\; 
\textbf{Bill Yuchen Lin}\; \;\;
\textbf{Radha Poovendran}\\
  University of Washington \\
  {  \texttt{\{zxu9,fqjiang,luyaoniu,byuchen,rp3\}@uw.edu}} \\
}

\begin{document}
\maketitle
\begin{abstract}


Instruction tuning has been widely adopted to ensure large language models (LLMs) follow user instructions effectively. 
The resulting instruction-following capabilities of LLMs heavily rely on the instruction datasets used for tuning.
Recently, synthetic dataset generation has emerged as an economically viable solution to provide LLMs diverse and high-quality instructions.
However, existing approaches typically assume that larger or stronger models are stronger teachers for instruction tuning, and hence adopt these models as response generators to distill knowledge for synthetic instructions.
In this paper, we challenge this commonly-adopted assumption.
Our extensive experiments across five base models and twenty response generators reveal that larger and stronger models are not necessarily stronger teachers of smaller models.
We refer to this phenomenon as the \emph{Larger Models' Paradox}.
We observe that existing metrics cannot precisely predict the effectiveness of response generators since they ignore the compatibility between teachers and base models being fine-tuned. We thus develop a novel metric, named as Compatibility-Adjusted Reward (CAR) to measure the effectiveness of response generators. 
Our experiments across five base models demonstrate that CAR outperforms almost all baselines. 
\end{abstract}

\section{Introduction}

\begin{figure}[!t]
    \centering
    \includegraphics[width=\linewidth]{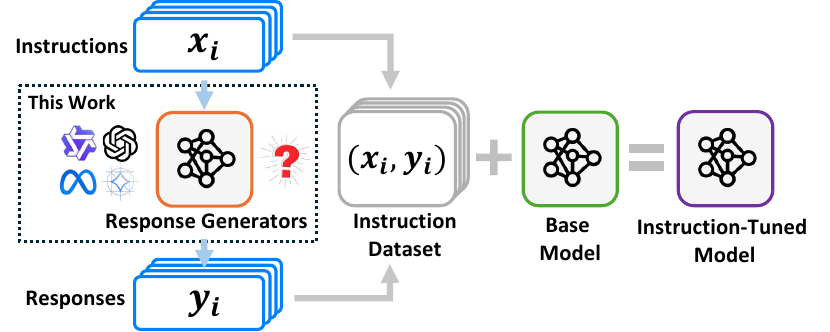}
    \caption{This figure demonstrates the process of instruction tuning and the scope of this paper.}
    \label{fig:scope}
\end{figure}

Instruction tuning (Figure \ref{fig:scope}) has been widely adopted to tailor the behavior of base Large Language Models (LLMs) to align with specific tasks and user intents \citep{zhang2023instruction}. This approach leverages instruction datasets, consisting of samples pairing an instruction with a corresponding response. The success of instruction tuning depends on the availability of high-quality instruction datasets. Initially, constructing these datasets required large human effort in generating and curating instruction-response pairs \citep{Dolly, zheng2024lmsyschatm, zhao2024wildchat}, which is time-consuming and labor-intensive \cite{liu2024best}.

To reduce the reliance on human-curated datasets, synthetic datasets generated by LLMs have surfaced as a viable solution \cite{adler2024nemotron}. Recent works, such as \citep{sun2024principle, alpaca, wang-etal-2023-self-instruct, xu2024magpie, chen2024genqa}, have shown the strong potential of synthetic datasets in instruction tuning. While current research has primarily focused on using LLMs to create large, diverse, and high-quality instructions \citep{liu2024best}, the selection of appropriate LLMs for generating corresponding responses remains largely unexplored. The common approach relies on distilling from state-of-the-art models that excel in benchmark evaluations \citep{open-llm-leaderboard-v2, chiang2024chatbot} to generate responses for instruction tuning.
For instance, \emph{Llama-3.2-3B-Instruct} uses responses generated by \emph{Llama-3.1-405B-Instruct} (i.e., the largest model in Llama-3.1 family) for instruction tuning \cite{llama32_webpage}. Additionally, most of the existing open synthetic datasets \cite{OpenHermes2.5, xu2023wizardlm, ding2023ultrachat, alpaca-gpt4, chen2024genqa} depend on expensive, closed-source models like \emph{GPT-4} \cite{achiam2023gpt4} and \emph{Gemini} \citep{gemini} to produce responses.

\textit{Is it always better to use the larger or stronger models as teachers?} In this paper, we investigate the choice of the teacher model that generate responses during synthetic dataset generation, which we refer to as \textbf{response generators}, influence the instruction-following performance of the instruction-tuned LLMs. Specifically, given a base model and a set of high-quality instructions, we investigate the following research questions: 

\textbf{RQ1:} \textit{Which models are the most effective response generators for instruction tuning?} 

To answer RQ1, we conduct extensive experiments with five base models, and fine-tune them on datasets generated by 20 response generators across seven model families: Qwen2, Qwen2.5, Llama 3, Llama 3.1, Gemma 2, Phi-3, and GPT-4. Our findings challenge common assumptions in the field, revealing a surprising result which we term the \textbf{Larger Models' Paradox}: larger response generators (e.g., \emph{Llama-3.1-405B-Instruct}) do not always enhance a base model's instruction-following capabilities compared to their smaller counterparts within the same model family  (e.g. \emph{Llama-3.1-70B-Instruct}). Moreover, we find that open-source models (e.g., \emph{Gemma-2-9b-it} and \emph{Qwen2.5-72B-Instruct}) outperform \emph{GPT-4} as response generators. These findings question established practices and suggest more efficient and accessible approaches to create high-quality instruction datasets.

To further explore the Larger Models' Paradox, we investigate statistical metrics to reveal potential factors influencing the effectiveness of response generators. Here, we pose our second research question: 

\textbf{RQ2:} \textit{How can we determine the most effective response generators for a certain base model without instruction tuning?}

This question is crucial due to the significant computational costs associated with instruction tuning across multiple datasets generated by diverse response generators. Our investigation reveals that existing metrics in alignment data selection, including quality \citep{dubey2024llama}, difficulty \citep{li2023quantity}, and response length \citep{liu2023makes}, fail to consider the \textbf{compatibility} between the base model being fine-tuned and the response generator, thus results in their inability to explain the Larger Models' Paradox. To bridge this gap, we formulate the task of finding the most effective response generators as a risk-return problem. We solve this by calculating an \textbf{Compatibility-Adjusted Reward (CAR)}, where compatibility serves as the risk factor. This compatibility is quantified by the average loss of responses on the base model being fine-tuned, with higher average loss indicating lower compatibility and thus higher risk.
Our comparison of the proposed CAR with existing metrics demonstrates that it outperforms all baselines in predicting the effectiveness of response generators.

We believe that our findings on the Larger Models' Paradox and the proposed CAR can effectively guide future instruction tuning of LLMs. Instead of selecting response generators solely based on benchmark performance (e.g., GPT-4), practitioners should prioritize those with higher compatibility to better enhance the instruction-following capabilities of their LLMs.



\section{Related Work}

\paragraph{Synthetic Data Generation for Instruction Tuning.}

While human-crafted instruction datasets \citep{Dolly, zheng2024lmsyschatm, zhao2024wildchat} have been used for LLM instruction tuning, they are time-consuming and labor-intensive. Consequently, synthetic dataset generation has emerged as a promising alternative. Early approaches \citep{wang-etal-2023-self-instruct, alpaca, xu2023wizardlm, xu-etal-2023-baize, wang2024codeclm, luo2023wizardcoder, sun2024principle} focused on prompting LLMs to generate synthetic instructions, starting with a small set of human-annotated seed instructions and expanding these through few-shot prompting \citep{li2024synthetic}. Another line of work \citep{ding2023ultrachat, li2024synthetic} summarized world knowledge to generate more diverse synthetic datasets. Recent advancements \citep{xu2024magpie, chen2024genqa} further simplified the process by leveraging single prompts to sample instructions directly from LLMs, requiring minimal human oversight. While existing work primarily focused on generating large, diverse, and high-quality instructions, the impact of response generators is often overlooked.

\paragraph{Metrics for Data Selection.}
Instruction tuning data selection involves determining which instruction-response pairs to be included in the training dataset and how to sample them \citep{albalak2024survey}. The most widely-used metric for selecting instruction data is quality, which is often assessed using LLM evaluators \cite{chen2023alpagasus, liu2024selectit}, reward models \citep{dubey2024llama, xu2024magpie}, gradient similarity search \citep{xia2024less}, or a combination of these methods \cite{cao2024instruction}. Another key metric is difficulty, where higher difficulty is considered more valuable for learning. For instance, \citet{li2023quantity} introduces IFD, which measures the instruction-following difficulty of specific instruction-response pairs. \citet{li2024superfiltering} further refines IFD by utilizing GPT-2 for efficient estimation. Approaches like Deita \citep{liu2023makes} consider both quality and difficulty when selecting datasets. Token length is also adopted as a metric, as discussed in \citep{xia2024rethinkingdataselectionscale, liu2023makes}. Selective Reflection-Tuning \citet{li2024selective} approach selects and refines existing instruction-following datasets to address the inconsistency between teacher and student models. 

Our investigation complements existing research on alignment data selection by shifting the focus to the response generation process itself, as illustrated in Figure \ref{fig:scope}. While prior studies have concentrated on selecting the most effective instruction-response pairs with an existing instruction dataset, we explore the crucial role that response generators play in influencing the quality of instruction tuning.
\section{Which Models are the most effective teachers for instruction tuning?}
\label{sec: empirical}
\subsection{Preliminaries}

\paragraph{Instruction Datasets.} 
An instruction dataset can be represented as $\mathcal{D} = {(x_i, y_i)}_{i=1}^{|\mathcal{D}|}$, where each sample $(x_i,y_i)$ consists of an instruction $x_i$ and its corresponding response $y_i$. 
In this paper, we investigate how the response generator, denoted as $\mathcal{M}$, impacts the instruction-following capabilities of models fined-tuned with $\mathcal{D}$ with $y_i = \mathcal{M}(x_i)$.

\paragraph{Supervised Fine-Tuning.}

Supervised fine-tuning (SFT) is widely adopted to enhance instruction-following capabilities of LLMs. The SFT updates the parameters $\theta$ of a pre-trained language model to minimize the negative log-likelihood loss over the instruction dataset $\mathcal{D}$. 
The SFT loss can be formally expressed as:
\begin{equation}
\mathcal{L}_{\text{SFT}}(\theta) = - \frac{1}{|\mathcal{D}|}\sum_{(x_i, y_i) \in \mathcal{D}} \log p_{\theta}(y_i|x_i).
\end{equation}

\subsection{Experimental Setup}

\paragraph{Instruction Sets.}
To construct diverse and high-quality instructions, we sample from the Magpie-Air-3M dataset \citep{xu2024magpie}, and obtain a subset of 100K high-quality instructions, denoted as \textbf{Magpie-100K}. A detailed categorization of instruction tasks is provided in Appendix \ref{appendix: instruction data details}. 
Additionally, we extracted another 100K high-quality instructions from multiple sources, including UltraFeedback \citep{cui2023ultrafeedback}, WildChat \citep{zhao2024wildchat}, Lmsys-Chat-1M \citep{zheng2024lmsyschatm}, and Alpaca-GPT-4 \citep{alpaca-gpt4}. This instruction set, denoted as \textbf{Mix-100K}, contains both human-written and synthetic instructions, ensuring a comprehensive representation of instruction types.

\paragraph{Response Generators.} Our study considers 20 response generators across 7 model families for response generation. The model families include Qwen2 \citep{qwen2}, Qwen2.5 \citep{qwen2.5}, Llama 3 \citep{llama31}, Llama 3.1 \citep{llama31}, Gemma 2 \cite{team2024gemma}, Phi-3 \cite{abdin2024phi}, and GPT-4 \cite{achiam2023gpt4}. A comprehensive overview of the response generators is presented in Table \ref{tab:model_generators}. By combining the instructions with corresponding responses generated by these teacher models, we construct instruction-response pairs for instruction-tuning. By default, we use greedy decoding to generate responses. The datasets used in our experiments can be found here\footnote{https://huggingface.co/datasets/Magpie-Align/Magpie-100K-Generator-Zoo}.

\begin{table}[tbp]
\centering
\caption{Overview of 20 response generators used in our study.}
\resizebox{\columnwidth}{!}{%
\begin{tabular}{cc c r}
\toprule
\textbf{Model Family} & \textbf{Release Date} & \textbf{Model ID} & \bf Size \\
\midrule
\multirow{3}{*}{\makecell{\textbf{Qwen2}\\\cite{qwen2}}} & \multirow{3}{*}{Jun, 2024} &  Qwen2-1.5B-Instruct & 1.5B\\
& &  Qwen2-7B-Instruct & 7B \\
& &  Qwen2-72B-Instruct & 72B \\
\midrule
\multirow{5}{*}{\makecell{\textbf{Qwen2.5}\\ \cite{qwen2.5}}} & \multirow{5}{*}{Sept, 2024} & Qwen2.5-3B-Instruct & 3B \\
& &  Qwen2.5-7B-Instruct & 7B \\
& &  Qwen2.5-14B-Instruct & 14B \\
& &  Qwen2.5-32B-Instruct & 32B \\
& &  Qwen2.5-72B-Instruct & 72B \\
\midrule
\multirow{2}{*}{\makecell{\textbf{Llama 3}\\ \citep{llama31}}} & \multirow{2}{*}{Apr, 2024} &  Llama-3-8B-Instruct & 8B\\
& &  Llama-3-70B-Instruct & 70B \\
\midrule
\multirow{3}{*}{\makecell{\textbf{Llama 3.1}\\ \citep{llama31}}} & \multirow{3}{*}{Jul, 2024} &  Llama-3.1-8B-Instruct & 8B \\
& &  Llama-3.1-70B-Instruct & 70B \\
& &  Llama-3.1-405B-Instruct & 405B \\
\midrule
\multirow{3}{*}{\makecell{\textbf{Gemma 2}\\ \cite{team2024gemma}}} & \multirow{3}{*}{Jun, 2024} &  Gemma-2-2b-it & 2B\\
& &  Gemma-2-9b-it & 9B \\
& &  Gemma-2-27b-it & 27B \\
\midrule
\multirow{3}{*}{\makecell{\textbf{Phi-3}\\ \cite{abdin2024phi}}} & \multirow{3}{*}{Jun, 2024} &  Phi-3-mini-128k-instruct & 3.8B \\
& &  Phi-3-small-128k-instruct & 7B \\
& &  Phi-3-medium-128k-instruct & 14B  \\
\midrule
\multirow{2}{*}{\makecell{\textbf{GPT-4}\\ \cite{achiam2023gpt4}}} & \multirow{2}{*}{\makecell{Since\\ Mar, 2023}} & \multirow{2}{*}{\makecell{GPT-4 \& GPT-4 Turbo} }& \multirow{2}{*}{-} \\
&  \\ \bottomrule
\end{tabular}%
}
\label{tab:model_generators}
\end{table}

\paragraph{Base Models.} We consider five base language models from different developers of varying sizes as students, including Qwen2-1.5B \citep{qwen2}, Gemma-2-2b \citep{team2024gemma}, Llama-3.2-3B \citep{llama32}, Qwen2.5-3B, \cite{qwen2.5} and Llama-3.1-Minitron-4B-Width-Base (Llama-3.1-Minitron-4B) \citep{minitron2024}.

\paragraph{Evaluation Benchmarks.} To evaluate the instruction-following capabilities of the instruction-tuned models, we use two widely-used instruction-following benchmarks: \textbf{AlpacaEval 2} (\textbf{AE2}) \citep{alpaca_eval} and \textbf{Arena-Hard} (\textbf{AH}) \citep{arenahard2024}.
Specifically, AE2 contains 805 representative instructions from real user interactions. AH contains 500 challenging user queries. 
AE2 and AH use \emph{GPT-4-Turbo (1106)} and \emph{GPT-4-0314} as the baselines to assess the performance of instruction-tuned models, respectively.
Both benchmarks compare responses generated by the model of interest with those generated by baselines, and employ GPT evaluators to automatically annotate which response is preferred.

\paragraph{Evaluation Metrics.} Similar to existing studies, we adopt two metrics to measure the performance of fine-tuned SLMs. The first metric, used by both benchmarks, is the \textbf{win rate (WR)}, which calculates the fraction of responses that are favored by the GPT evaluator. The second metric, used by AE2, is the \textbf{length-controlled win rate (LC)} \cite{dubois2024length}. LC accounts for response length to reduce its impact on WR. Additionally, we report the \textbf{Average Performance (AP)}, computed as the mean of AE2's LC and AH's WR.

\paragraph{Instruction-Tuning and Evaluation Setup.} 

We use SFT and implement a cosine learning rate schedule with a max learning rate of $2 \times 10^{-5}$ to fine-tuning the base models for 2 epoches \citep{touvron2023llama}. The detailed hyper-parameters and experimental platform can be found in Appendix \ref{appendix: Instruction-Tuning Setups}.
We follow the official instruction templates of each model. To ensure reproducibility of our empirical analysis, we implement greedy decoding for both AE2 and AH benchmarks.

\subsection{Empirical Evaluation}
\begin{figure}[!t]
    \centering
    \includegraphics[width=\linewidth]{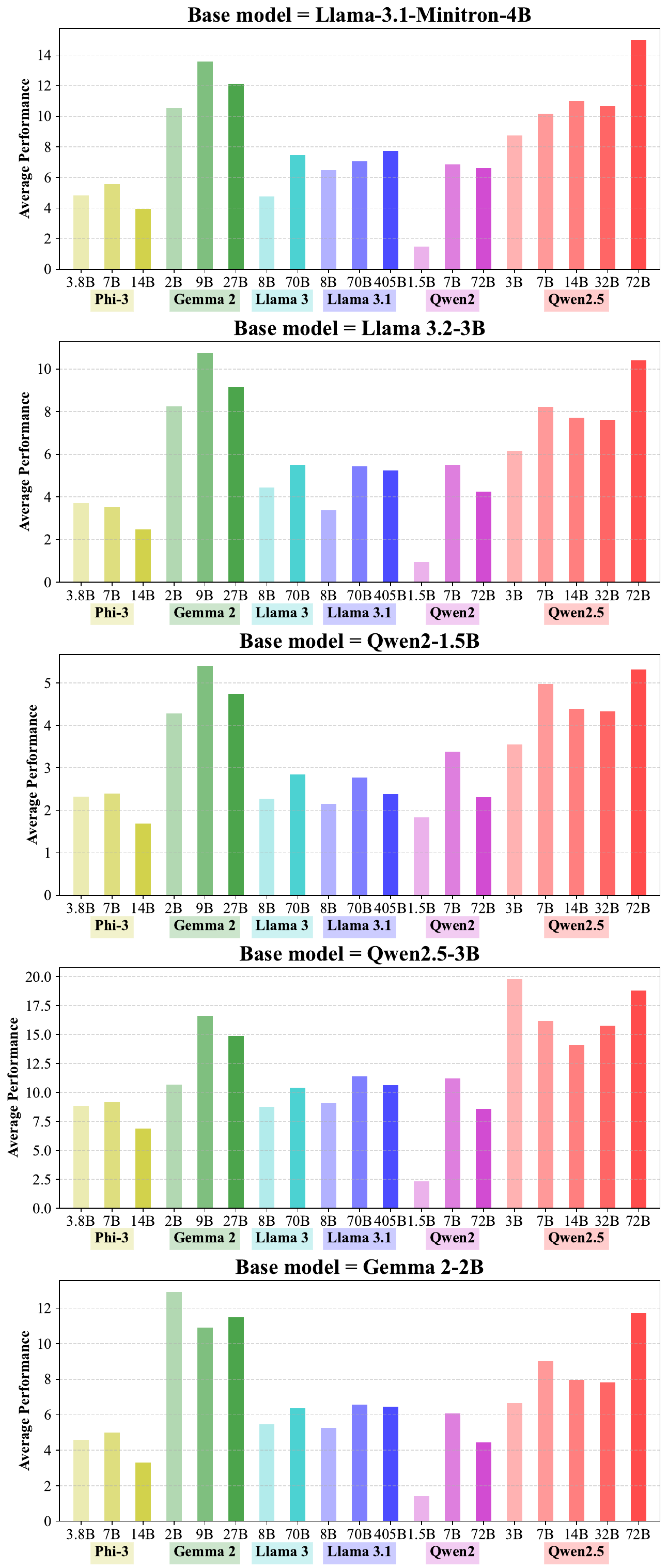}
    \caption{Average performance of five base models fine-tuned on various response generators across six model families. We use different colors to distinguish between model families, with darker bars indicating larger response generators within each family.}
    \vspace{-2em}
    \label{fig:main-base-response-eval}
\end{figure}


This section evaluates the instruction-following capabilities of models fine-tuned over datasets whose responses are generated by various response generators.
By default, we utilize the Magpie-100K dataset as our primary instruction set. Figure \ref{fig:main-base-response-eval} provides a comprehensive overview of the AP across different base models and response generators, and the detailed benchmark scores of AE2 and AH are deferred to Table \ref{table: full results} in Appendix \ref{appendix: detailed benchmark scores}.
\textcolor{black}{Evaluations on larger base model (Llama-3.1-8B) with different response generators are presented in Table \ref{tab: large base models} in Appendix \ref{appendix: larger base model}.
We analyze the effect of data randomness on average performance in Table \ref{tab: data randomness}.
}

We observe that the Gemma-2 and Qwen2 families consistently demonstrate superior performance across all base models evaluated. Notably, \textbf{Gemma-2-9b-it} and \textbf{Qwen2.5-72B-Instruct} emerge as the two best response generators, as evidenced by their consistently high AP scores.
In addition, we report the following key findings.

\begin{findingBox}{1}{
\textbf{[Larger Models' Paradox]} Larger response generators $\centernot\implies$ improved instruction-following capabilities.
}
\end{findingBox}

Our evaluation reveals a \emph{counterintuitive finding}: increasing the model size of response generators does not necessarily improve the instruction-following capabilities of base models \textcolor{black}{within the same model family}.
This finding is universal, evidenced across multiple model families. 
For example, \emph{Gemma-2-9b-it} demonstrates superior performance compared to its larger counterpart, \emph{Gemma-2-27b-it}, in SFT across almost all base models examined. 
Similar observations are made in other model pairs: \emph{Phi-3-Small} outperforms \emph{Phi-3-Medium}, \emph{Llama-3.1-70B-Instruct} surpasses \emph{Llama-3.1-405B-Instruct}, \emph{Qwen2-7B-Instruct} outperforms \emph{Qwen2-72B-Instruct}, and \emph{Qwen2.5-7B-Instruct} exceeds \emph{Qwen2.5-32B-Instruct}. 
We refer to this finding as the \textbf{Larger Models Paradox}: larger language models, despite their superior performance, may not always generate better responses for fine-tuning smaller language models \textcolor{black}{within the same model family} compared to responses generated by medium-sized models.

We believe the key to explain this paradox is the \textbf{compatibility} between the response generators and base models. 
For example, a high-quality textbook (responses from large size response generators) written for college students may be challenging for primary school students (smaller base models). 
We will investigate this paradox in Section \ref{sec: metric} with more detailed statistics and metrics to evaluate the compatibility.

\begin{findingBox}{2}{
\textbf{[Family's Help]} Learning from response generators within the same model family leads to higher performance.
}
\end{findingBox}

We observe higher AP when base models are fine-tuned using responses generated by models within the same family. This is evidenced when Qwen2-1.5B, Qwen2.5-3B, and Gemma 2-2B serve as base models. In these instances, the relative performance of using intra-family response generators surpasses that observed when tuning other base models.

Furthermore, while not practically applicable, we observe a significant performance boost when fine-tuning a base model using responses generated from its own instruction-tuned version. A prime example of this is the Gemma 2-2B base model, which achieves best performance when tuned with responses from \emph{Gemma-2-2b-it}, outperforming all other response generators. These two phenomena underscore the importance of compatibility between the base model and the response generator in instruction tuning.

\begin{findingBox}{3}{
\textbf{[Open-Source $>$ Close-Source]} Open-source LLMs can outperform close-source LLMs as response generators.
}
\end{findingBox}


Table \ref{tab: gpt_vs_others} compares the instruction-tuning performance when utilizing GPT-4 and open-source LLMs (e.g., Gemma 2, Llama 3, Llama 3.1 and Qwen2.5) as response generators. For this evaluation, we employ the Mix-100K dataset as our instruction source. Notably, our findings reveal that all open-source LLMs significantly outperform GPT-4. 
\textcolor{black}{We hypothesize that this is because the response length of GPT-4 is less than open-source LLMs, thus less favored by the evaluators.}
These results suggest the potential for using cost-effective open-source LLMs for synthetic data generation in instruction-tuning tasks.

\begin{table}[tbp]
    \centering
    \caption{This table compares the performance of GPT-4 and other state-of-the-art open source LLMs as the response generator. All models are supervised-fine-tuned on the Llama-3.1-Minitron-4B base model.}
    \resizebox{\columnwidth}{!}{%
    \begin{tabular}{c c c c c}\toprule
        \bf Response  & \multicolumn{2}{c}{\textbf{AlpacaEval 2}} & \textbf{Arena-Hard} & \textbf{AP} \\
       \bf Generator Model & LC (\%) & WR (\%) & WR (\%) &  (\%) \\ \midrule 
       Gemma-2-9b-it & 16.09 & 13.70 & 13.7 & 14.90 \\
       Gemma-2-27b-it & 13.93 & 13.31 & 12.4 & 13.17 \\
       Llama-3-70b-Instruct & 10.55 & 10.68 & 6.7 & 8.62 \\
       Llama-3.1-70b-Instruct & 9.52 & 10.10 & 8.3 & 8.91 \\
       Qwen2.5-7B-Instruct & 13.50 & 14.33 & 10.6 & 12.05 \\
       Qwen2.5-72B-Instruct & \textbf{19.20} & \textbf{21.01} & \textbf{13.1} & \textbf{16.15} \\ \midrule
       GPT-4 & 6.63 & 5.70 & 4.8 & 5.72 \\
        \bottomrule
    \end{tabular}
    }
    \vspace{-1em}
    \label{tab: gpt_vs_others}
\end{table}

\begin{findingBox}{4}{
Higher temperature and top-p enhance instruction-following capabilities.
}
\end{findingBox}

\begin{figure}[htbp]
    \centering
    \includegraphics[width=0.8\linewidth]{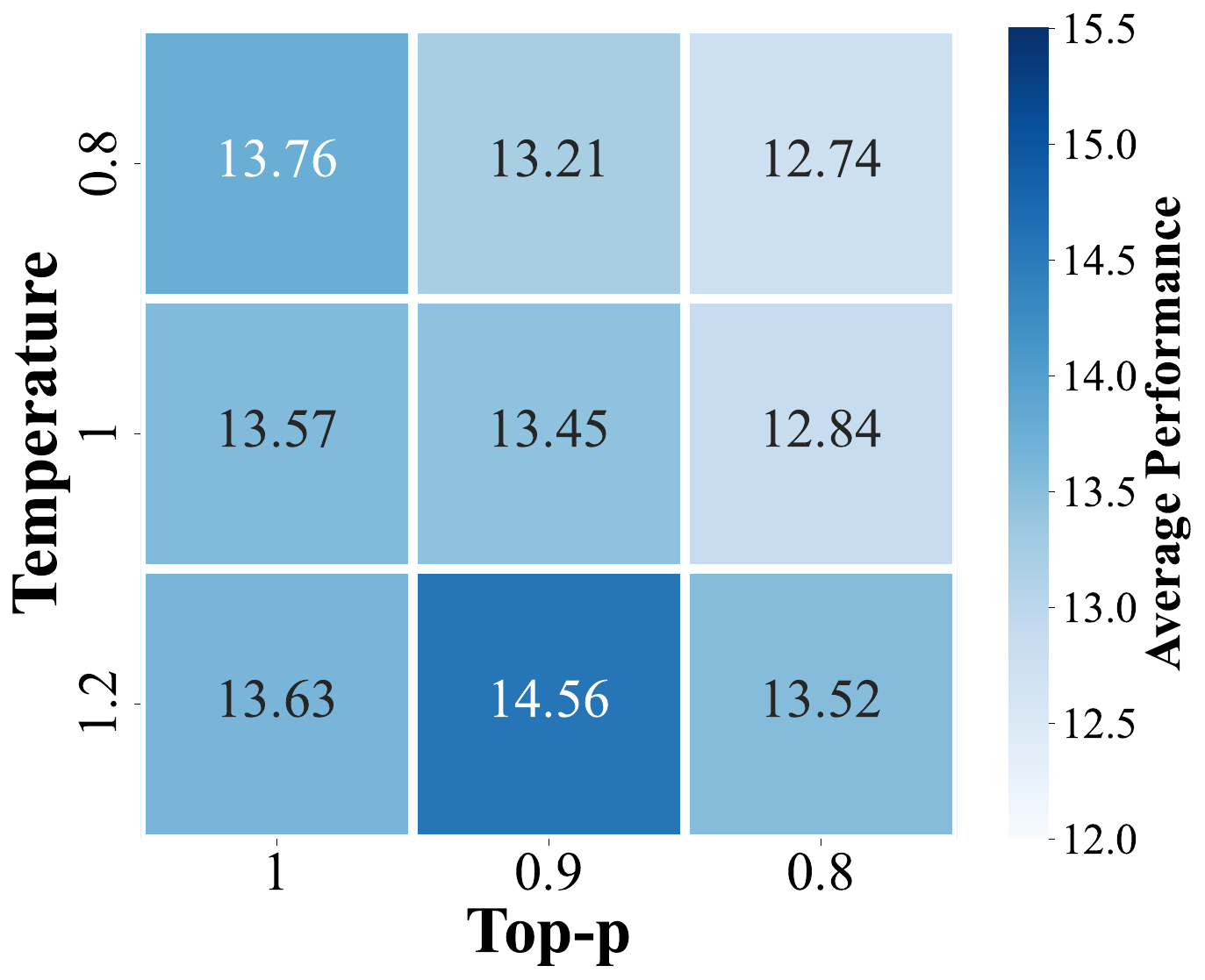}
    \caption{This figure demonstrates the impact of different sampling hyper-parameters when generating responses. We use \emph{Gemma-2-9b-it} as the response generator. All models are supervised-fine-tuned on the Llama-3.1-Minitron-4B base model.}
    \label{fig:sampling_params}
\end{figure}

Figure \ref{fig:sampling_params} illustrates the effects of different sampling hyper-parameters when generating responses using \emph{Gemma-2-9b-it} model. We observe that higher temperature and top-p value can lead to better performance in instruction following. We hypothesize that this enhancement in performance is because higher temperature and top-p values yield more diverse and contextually rich outputs.

\begin{findingBox}{5}{
Reject sampling slightly increases instruction-tuning performance.
}
\end{findingBox}

Table \ref{tab: reject sampling} quantifies the impact of reject sampling on synthetic data generation using \emph{Gemma-2-9b-it} model. Specifically, we generate 5 responses per instruction with temperature $T=0.8$, evaluate them using the \emph{ArmoRM-Llama3-8B-v0.1} reward model \cite{ArmoRM}, and select the highest and lowest-rated responses to create two distinct datasets: Best-of-N and Worst-of-N. We also compare them with responses sampled at $T=0.8$ and greedy decoding ($T=0$). The results presented in Table \ref{tab: reject sampling} demonstrate a slight improvement in performance when utilizing reject sampling compared to standard sampling techniques. 

\begin{table}[tbp]
\centering
\caption{This table investigates the impact of reject sampling on model performance.}
\resizebox{\columnwidth}{!}{%
\begin{tabular}{c c c c c c}\toprule
\multicolumn{1}{c}{\multirow{2}{*}{\textbf{Base Model}}} & 
\multicolumn{1}{c}{\multirow{2}{*}{\textbf{Method}}} &
\multicolumn{2}{c}{\textbf{AlpacaEval 2}} & 
\textbf{Arena-Hard} & \textbf{AP} \\
& & LC (\%) & WR (\%) & WR (\%) & (\%)  \\ \midrule
\multirow{4}{*}{\makecell{Llama-3.1-\\Minitron-4B}} & Best-of-N & \textbf{15.94} & \textbf{15.14} & \textbf{11.9} & \textbf{13.92} \\
& Worst-of-N & 13.02 & 12.66 & 11.0 & 12.01 \\
& Sampling & 15.71 & 14.81 & 11.8 & 13.755 \\
& Greedy & 16.13 & 14.51 & 11.0 & 13.565 \\ \midrule
\multirow{4}{*}{\makecell{Qwen2.5-\\3B-Instruct}} & Best-of-N & \textbf{13.83} & \textbf{13.57} & \textbf{21.0} & \textbf{17.415} \\
& Worst-of-N & 12.37 & 12.54 & 17.9 & 15.135 \\
& Sampling & 13.43 & 13.29 & 20.1 & 16.765 \\
& Greedy & 13.78 & 13.57 & 19.4 & 16.59 \\
\bottomrule
\end{tabular}
}
\label{tab: reject sampling}
\end{table}



In what follows, we summarize the conclusion for RQ1.

\begin{tcolorbox}[title=RQ1. Which models are the most effective response generators for instruction tuning?]
A1. Gemma-2 and Qwen2 families consistently demonstrate superior performance across all base models evaluated, and even \textbf{outperform GPT-4}. Notably, \textbf{Gemma-2-9b-it} and \textbf{Qwen2.5-72B-Instruct} emerge as the two best response generators, as evidenced by their consistently high AP scores. We also found that \textbf{larger models do not always generate responses for enhanced instruction-following capabilities}.
\end{tcolorbox}

\section{How can we determine the most effective response generators without instruction tuning?}
\label{sec: metric}

\subsection{Measure the Effectiveness of Response Generators}

It is computationally expensive to brute-force all response generators to identify the most effective one for a given base model. 
In this section, we investigate how to measure the effectiveness of response generators for a given base model without training or fine-tuning.
Specifically, we study the following research question:

\begin{definition}[Effectiveness Measure of Response Generators]
    Given a base language model and a set of synthetic instruction datasets ${\mathcal{D}_1, \mathcal{D}_2, ..., \mathcal{D}_n}$, where each $\mathcal{D}_i$ contains responses generated by a distinct response generator $\mathcal{M}_i$, measure the effectiveness of these response generators without performing the actual fine-tuning process.
\end{definition}

\paragraph{Evaluation Metric.}
To assess the accuracy when measuring effectiveness of response generators, we employ Spearman's rank correlation coefficient ($\rho$) \cite{zar2005spearman}. This coefficient evaluates the monotonic relationship between two ranking variables. In our context, we compute $\rho$ between two ranks: the ground truth rank $R_{AP}$, obtained by fine-tuning the model on each synthetic instruction dataset and measuring the Average Performance (AP), and an estimated rank $R_{EST}$, predicted without fine-tuning. Spearman's $\rho$ is calculated as:
\begin{equation}
\rho = 1 - \frac{6\sum d_i^2}{n(n^2 - 1)}
\end{equation}
where $d_i$ is the difference between the two ranks for each observation and $n$ is the number of observations. $\rho$ ranges from -1 to 1, with 1 indicating a perfect positive correlation. Our objective is to maximize $\rho$, thereby achieving the closest prediction between predicted and actual performance rankings. \textcolor{black}{We employ the empirical results obtained in Section \ref{sec: empirical} as the ground truth.}

\subsection{Baseline Methods}

In this section, we introduce commonly-used metrics for alignment data selection: quality, difficulty, and response length, for predicting the performance rank of instruction-tuned models. 

\paragraph{Response Quality.} Following \citet{llama32, xu2024magpie}, we assess response quality using reward models and calculate the \textbf{Average Reward (AR)} of all responses. To mitigate potential selection bias, we employ three state-of-the-art reward models from RewardBench \citep{RewardBench}: \emph{ArmoRM-Llama3-8B-v0.1} \citep{ArmoRM}, \emph{Skywork-Reward-Llama-3.1-8B} \citep{skyworkreward2024}, and \emph{Skywork-Reward-Gemma-2-27B} \citep{skyworkreward2024}.

\paragraph{Instruction-following Difficulty.} Instruction-following difficulty is another widely-used metric in alignment data selection \citep{llama32,liu2023makes,li2023quantity,li2024superfiltering,xu2024magpie}. To assess the difficulty of responses, we employ the following two metrics:
\begin{enumerate}
    \item \textbf{Response Perplexity (PPL).} For a given instruction-response pair $(x_i,y_i)$, the response perplexity is defined as:
    \begin{equation*}
    \begin{aligned}
    & \text{PPL}(y_i|x_i) = \\ & \exp(-\frac{1}{N} \sum_{j=1}^N \log 
     p_\theta(y_{i,j}|x_i, y_{i,1:j-1})),
    \end{aligned}
    \end{equation*}
    where $N$ is the token length of $y_i$ and $y_{i,j}$ is its $j$-th token, and $\theta$ is the parameter of the base model. We use \emph{GPT-2} model and each corresponding base model for evaluation, denoted as PPL-GPT2 and PPL-Self respectively.



\item \textbf{Instruction Following Difficulty (IFD) \citep{li2023quantity}.}  IFD is defined as:
\begin{equation*}
    \text{IFD}(y_i|x_i) = \frac{\text{PPL}(y_i|x_i)}{\text{PPL}(y_i)},
\end{equation*}
where $\text{PPL}(y_i)$ is the unconditional perplexity of response $y_i$. We follow \citet{li2024superfiltering} and employ \emph{GPT-2} and the base model respectively, denoted as IFD-GPT2 and IFD-Self.
\end{enumerate}

For each metric, we compute the average value across the entire dataset $\mathcal{D}_i$. 

\paragraph{Response Length.} According to \citet{liu2023makes} and \citet{xia2024rethinkingdataselectionscale}, the response length positively correlates with the final alignment performance. We use the \texttt{tiktoken} library \citep{tiktoken} to count the number of response tokens for each pair, and report the average response length for each $\mathcal{D}_i$.

\subsection{Baseline Methods Fails to Measure the Effectiveness of Response Generators}

In what follows, we demonstrate that the effectiveness of response generators indicated by baseline methods does not match the performance of models fine-tuned on various synthetic instruction datasets. 



\begin{figure}[htbp]
    \centering
    \includegraphics[width=1\linewidth]{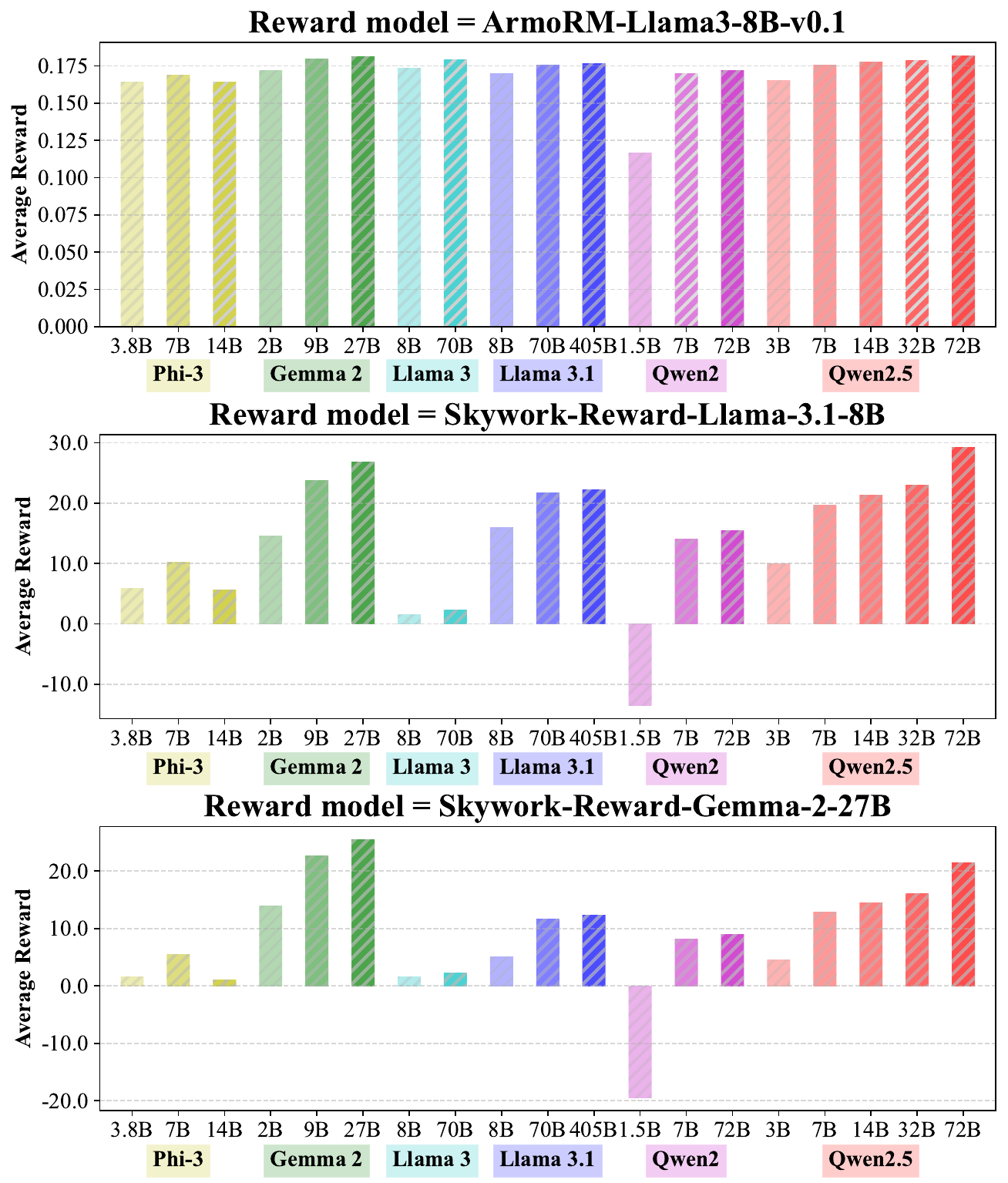}
    \caption{This figures demonstrates the response quality measured by three reward models.}
    \vspace{-1em}
    \label{fig:metric-reward}
\end{figure}

As shown in Figure \ref{fig:metric-reward}, AR consistently increases with model size within model families (except Phi-3 family). However, this trend fails to explain the "Larger Models Paradox" discussed in Section \ref{sec: empirical}. Notably, since AR measures human preference, this discrepancy suggests that responses preferred by humans are not necessarily optimal for aligning language models.

\begin{table*}[htbp]
    \vspace{1em}
    \centering
    \caption{Spearman's rank correlation coefficient ($\rho$) for different measurement metrics. Here $\mathcal{RM}_1$, $\mathcal{RM}_2$ , $\mathcal{RM}_3$  are reward models \emph{ArmoRM-Llama3-8B-v0.1}, \emph{Skywork-Reward-Llama-3.1-8B}, and \emph{Skywork-Reward-Gemma-2-27B}  respectively. We observe that our proposed CAR shows the highest correlation between the effectiveness of the response generator and the instruction-following capabilities of fine-tuned base models.}
    \resizebox{\textwidth}{!}{ 
    \begin{tabular}{c c  c  c c c c c c c }\toprule
     \multirow{2}[2]{*}{\bf Base Models}    & \multicolumn{3}{c}{\bf Reward} & \multicolumn{4}{c}{\bf Difficulty} & \multirow{2}[2]{*}{\shortstack{\bf Response \\ \bf Length}} &  \multirow{2}[2]{*}{\bf CAR} \\ \cmidrule(lr){2-4} \cmidrule(lr){5-8}
         & $\mathcal{RM}_1$ & $\mathcal{RM}_2$  & $\mathcal{RM}_3$  & IFD-GPT2 & IFD-Self & PPL-GPT2 & PPL-Self \\ \midrule
         
   \bf Qwen2-1.5B & 0.5526 & 0.7895 & 0.8754 & 0.7088 & 0.7719 & 0.1473 & 0.5596 & 0.5404 & \textbf{0.8842} \\
   \bf Gemma 2-2B & 0.5526 & 0.7982 & 0.8842 & 0.8281 & 0.8930 & 0.1614 & 0.4351 & 0.6298 & \textbf{0.9000} \\
   \bf Qwen2.5-3B & 0.4526 & 0.7351 & 0.7456 & 0.7386 & 0.8088 & 0.0456 & -0.0614 & 0.6088 & \textbf{0.8105} \\
   \bf Llama 3.2-3B & 0.6088 & 0.8105 & \textbf{0.9088} & 0.7632 & 0.8579 & 0.0456 & 0.6018 & 0.5877 & 0.9053 \\ 
   \bf Llama-3.1-Minitron-4B & 0.6632 & 0.8860 & 0.9386 & 0.7491 & 0.8555 & 0.1579 & 0.6263 & 0.5807 & \textbf{0.9439} \\ \midrule
  \bf  Average & 0.5660 & 0.8039 & 0.8705 & 0.7575 & 0.8374 & 0.1116 & 0.4323 & 0.5895 & \textbf{0.8888} \\ \bottomrule
    \end{tabular}
    }
    \label{tab: spearman}
\end{table*}

Similarly, metrics representing instruction-following difficulty (IFD and Perplexity) and response length show no strong correlation with model instruction-following capabilities. We deferred the results and analysis of these metrics to Appendix \ref{appendix: more baseline visualization}.
These findings highlight the inadequacy of existing metrics in accurately measuring the effectiveness of response generators in enhancing performance of instruction-tuned models.

\subsection{A Compatibility-Aware Metric to Measure Effectiveness}

In this section, we present a new metric to measure the effectiveness of response generators, making the "Larger Models Paradox" explainable. Our key insight to capture the \textbf{compatibility of response generators with base models}. To reflect such compatibility, we use the loss of the response $r_i$ in the base model being fine-tuned as the key metric. Intuitively, a lower loss of response $y_i$ on the base model indicates that the response aligns well with the base model's existing knowledge and capabilities, thus is more learnable compared to the response with higher loss. 

\textcolor{black}{While compatibility is crucial, it alone cannot fully measure effectiveness. Consider a scenario where a response generator consistently produces simple, low-quality responses for every question. In such cases, although these responses might be highly compatible with the base model, their overall quality and would be low.}
Therefore, to bridge this gap between quality and compatibility, we formulate the task of finding the most effective response generator as a risk-return problem \cite{fama1973risk}. We propose an adjusted reward value that incorporates both the potential benefit (return) and the compatibility risk. Specifically, we define our \textbf{Compatibility-Adjusted Reward (CAR)} as follows:
\begin{equation}
\text{CAR}(\mathcal{D}_i, \theta) = \frac{r(\mathcal{D}_i)}{1 + \beta \cdot {L}(\mathcal{D}_i, \theta)}
\end{equation}
where $r(\mathcal{D}_i)$ is the average reward measured by the reward model, representing the potential return, and ${L}(\mathcal{D}_i, \theta) = - \frac{1}{|\mathcal{D}_i|}\sum_{y_i \in \mathcal{D}_i} \log p_{\theta}(y_i)$ is the average loss for responses in $\mathcal{D}_i$ on the base model parameterized by $\theta$. $\beta$ is a tunable parameter that controls the impact of compatibility on the adjusted reward. 
\textcolor{black}{CAR penalizes the average reward from the reward model with the compatibility risk measured by the loss. This balanced approach enables quantitative assessment of the trade-off between the response quality and compatibility.}

\subsection{Experimental Results}

Table \ref{tab: spearman} compares the Spearman's $\rho$ correlation coefficient of baseline metrics with our CAR when using datasets generated by different response generators to fine-tune various base models. For CAR calculation, we employ \emph{Skywork-Reward-Gemma-2-27B} as the reward model and set $\beta=3$. The results in Table \ref{tab: spearman} demonstrate that our proposed CAR consistently outperforms other baseline metrics across almost all settings, indicating its potential to predict the effectiveness of different response generators without instruction tuning.

\begin{tcolorbox}[title=RQ2. How can we determine the most effective response generators without instruction tuning?]
A2. Existing metrics in instruction data selection are inadequate for accurate prediction as they fail to consider the compatibility between the base model and the response generator. To address this limitation, we propose the Compatibility-Adjusted Reward (CAR), which achieves better performance in identifying effective response generators across various base models.
\end{tcolorbox}
\section{Conclusion and Future Work}

This paper investigates the impact of response generators in synthetic dataset generation for instruction tuning. We uncovered the Larger Models' Paradox, wherein larger response generators do not necessarily enhance a base model's instruction-following capabilities compared to their smaller counterparts within the same model family. To explain this phenomenon, we considered the compatibility between response generators and the base model, and proposed the Compatibility-Adjusted Reward (CAR). Our metric achieved better performance in identifying the effectiveness of different response generators without the need for fine-tuning, outperforming existing baselines in alignment dataset selection.

We will explore several promising directions. First, efficiently transforming existing datasets to achieve better compatibility can lead to more effective use of available instruction tuning datasets. Second, investigating theoretical foundations of compatibility would enhance our understanding of the underlying mechanisms of instruction tuning. Lastly, studying the impact of different response generators for preference tuning may help aligning LLMs to better reflect human values.
\section*{Limitations}

While our study provides valuable insights into the effectiveness of response generators in instruction tuning, we acknowledge that our research primarily focuses on general instruction following tasks and does not extensively explore the synthesis of alignment datasets for specialized domains such as mathematics or complex reasoning. As a result, the applicability of the Larger Models' Paradox to these specific areas remains uncertain.

\section*{Ethical Impact}
This paper makes a counterintuitive observation, referred to as the Larger Models' Paradox, showing that stronger models are not stronger teachers for instruction tuning. 
We further propose a new metric to measure the effectiveness of teachers when generating responses for instruction datasets.
This metric informs the selection of response generators for future fine-tuning processes to enhance language models' instruction-following capabilities.
We do not identify potential misuse and ethical concerns in this paper.

\section*{Acknowledgment}

This work is partially supported by the Air Force Office of Scientific Research (AFOSR) under grant FA9550-23-1-0208, the National Science Foundation (NSF) under grants IIS 2229876, and the Office of Naval Research under grant N0014-23-1-2386.

This work is supported in part by funds provided by the National Science Foundation, by the Department of Homeland Security, and by IBM. Any opinions, findings, and conclusions or recommendations expressed in this material are those of the author(s) and do not necessarily reflect the views of the National Science Foundation or its federal agency and industry partners.

\bibliography{custom}

\begin{thebibliography}{53}
\providecommand{\natexlab}[1]{#1}

\bibitem[{Abdin et~al.(2024)Abdin, Jacobs, Awan, Aneja, Awadallah, Awadalla,
  Bach, Bahree, Bakhtiari, Behl et~al.}]{abdin2024phi}
Marah Abdin, Sam~Ade Jacobs, Ammar~Ahmad Awan, Jyoti Aneja, Ahmed Awadallah,
  Hany Awadalla, Nguyen Bach, Amit Bahree, Arash Bakhtiari, Harkirat Behl,
  et~al. 2024.
\newblock Phi-3 technical report: A highly capable language model locally on
  your phone.
\newblock \emph{arXiv preprint arXiv:2404.14219}.

\bibitem[{Achiam et~al.(2023)Achiam, Adler, Agarwal, Ahmad, Akkaya, Aleman,
  Almeida, Altenschmidt, Altman, Anadkat et~al.}]{achiam2023gpt4}
Josh Achiam, Steven Adler, Sandhini Agarwal, Lama Ahmad, Ilge Akkaya,
  Florencia~Leoni Aleman, Diogo Almeida, Janko Altenschmidt, Sam Altman,
  Shyamal Anadkat, et~al. 2023.
\newblock Gpt-4 technical report.
\newblock \emph{arXiv preprint arXiv:2303.08774}.

\bibitem[{Adler et~al.(2024)Adler, Agarwal, Aithal, Anh, Bhattacharya, Brundyn,
  Casper, Catanzaro, Clay, Cohen et~al.}]{adler2024nemotron}
Bo~Adler, Niket Agarwal, Ashwath Aithal, Dong~H Anh, Pallab Bhattacharya,
  Annika Brundyn, Jared Casper, Bryan Catanzaro, Sharon Clay, Jonathan Cohen,
  et~al. 2024.
\newblock Nemotron-4 340b technical report.
\newblock \emph{arXiv preprint arXiv:2406.11704}.

\bibitem[{Albalak et~al.(2024)Albalak, Elazar, Xie, Longpre, Lambert, Wang,
  Muennighoff, Hou, Pan, Jeong et~al.}]{albalak2024survey}
Alon Albalak, Yanai Elazar, Sang~Michael Xie, Shayne Longpre, Nathan Lambert,
  Xinyi Wang, Niklas Muennighoff, Bairu Hou, Liangming Pan, Haewon Jeong,
  et~al. 2024.
\newblock A survey on data selection for language models.
\newblock \emph{arXiv preprint arXiv:2402.16827}.

\bibitem[{Cao et~al.(2024)Cao, Kang, Wang, and Sun}]{cao2024instruction}
Yihan Cao, Yanbin Kang, Chi Wang, and Lichao Sun. 2024.
\newblock \href {https://openreview.net/forum?id=wF6k0aWjAu} {Instruction
  mining: Instruction data selection for tuning large language models}.
\newblock In \emph{First Conference on Language Modeling}.

\bibitem[{Chen et~al.(2024)Chen, Qadri, Wen, Jain, Kirchenbauer, Zhou, and
  Goldstein}]{chen2024genqa}
Jiuhai Chen, Rifaa Qadri, Yuxin Wen, Neel Jain, John Kirchenbauer, Tianyi Zhou,
  and Tom Goldstein. 2024.
\newblock Genqa: Generating millions of instructions from a handful of prompts.
\newblock \emph{arXiv preprint arXiv:2406.10323}.

\bibitem[{Chen et~al.(2023)Chen, Li, Yan, Wang, Gunaratna, Yadav, Tang,
  Srinivasan, Zhou, Huang et~al.}]{chen2023alpagasus}
Lichang Chen, Shiyang Li, Jun Yan, Hai Wang, Kalpa Gunaratna, Vikas Yadav,
  Zheng Tang, Vijay Srinivasan, Tianyi Zhou, Heng Huang, et~al. 2023.
\newblock Alpagasus: Training a better alpaca with fewer data.
\newblock \emph{arXiv preprint arXiv:2307.08701}.

\bibitem[{Chiang et~al.(2024)Chiang, Zheng, Sheng, Angelopoulos, Li, Li, Zhang,
  Zhu, Jordan, Gonzalez, and Stoica}]{chiang2024chatbot}
Wei-Lin Chiang, Lianmin Zheng, Ying Sheng, Anastasios~Nikolas Angelopoulos,
  Tianle Li, Dacheng Li, Hao Zhang, Banghua Zhu, Michael Jordan, Joseph~E.
  Gonzalez, and Ion Stoica. 2024.
\newblock \href {https://arxiv.org/abs/2403.04132} {Chatbot arena: An open
  platform for evaluating llms by human preference}.
\newblock \emph{Preprint}, arXiv:2403.04132.

\bibitem[{Cui et~al.(2023)Cui, Yuan, Ding, Yao, Zhu, Ni, Xie, Liu, and
  Sun}]{cui2023ultrafeedback}
Ganqu Cui, Lifan Yuan, Ning Ding, Guanming Yao, Wei Zhu, Yuan Ni, Guotong Xie,
  Zhiyuan Liu, and Maosong Sun. 2023.
\newblock Ultrafeedback: Boosting language models with high-quality feedback.
\newblock \emph{arXiv preprint arXiv:2310.01377}.

\bibitem[{Databricks(2023)}]{Dolly}
Databricks. 2023.
\newblock \href
  {https://huggingface.co/datasets/databricks/databricks-dolly-15k} {Databricks
  dolly-15k}.

\bibitem[{Ding et~al.(2023)Ding, Chen, Xu, Qin, Zheng, Hu, Liu, Sun, and
  Zhou}]{ding2023ultrachat}
Ning Ding, Yulin Chen, Bokai Xu, Yujia Qin, Zhi Zheng, Shengding Hu, Zhiyuan
  Liu, Maosong Sun, and Bowen Zhou. 2023.
\newblock Enhancing chat language models by scaling high-quality instructional
  conversations.
\newblock \emph{arXiv preprint arXiv:2305.14233}.

\bibitem[{Dubey et~al.(2024)Dubey, Jauhri, Pandey, Kadian, Al-Dahle, Letman,
  Mathur, Schelten, Yang, Fan et~al.}]{dubey2024llama}
Abhimanyu Dubey, Abhinav Jauhri, Abhinav Pandey, Abhishek Kadian, Ahmad
  Al-Dahle, Aiesha Letman, Akhil Mathur, Alan Schelten, Amy Yang, Angela Fan,
  et~al. 2024.
\newblock The llama 3 herd of models.
\newblock \emph{arXiv preprint arXiv:2407.21783}.

\bibitem[{Dubois et~al.(2024)Dubois, Galambosi, Liang, and
  Hashimoto}]{dubois2024length}
Yann Dubois, Bal{\'a}zs Galambosi, Percy Liang, and Tatsunori~B Hashimoto.
  2024.
\newblock Length-controlled alpacaeval: A simple way to debias automatic
  evaluators.
\newblock \emph{arXiv preprint arXiv:2404.04475}.

\bibitem[{Fama and MacBeth(1973)}]{fama1973risk}
Eugene~F Fama and James~D MacBeth. 1973.
\newblock Risk, return, and equilibrium: Empirical tests.
\newblock \emph{Journal of political economy}, 81(3):607--636.

\bibitem[{Fourrier et~al.(2024)Fourrier, Habib, Lozovskaya, Szafer, and
  Wolf}]{open-llm-leaderboard-v2}
Clémentine Fourrier, Nathan Habib, Alina Lozovskaya, Konrad Szafer, and Thomas
  Wolf. 2024.
\newblock Open llm leaderboard v2.
\newblock
  \url{https://huggingface.co/spaces/open-llm-leaderboard/open_llm_leaderboard}.

\bibitem[{Gallego(2023)}]{alpaca-gpt4}
Victor Gallego. 2023.
\newblock \href {https://huggingface.co/datasets/vicgalle/alpaca-gpt4}
  {alpaca-gpt4}.

\bibitem[{Google(2024)}]{gemini}
Google. 2024.
\newblock \href
  {https://blog.google/technology/ai/google-gemini-next-generation-model-february-2024/}
  {Our next-generation model: Gemini 1.5}.

\bibitem[{Lambert et~al.(2024)Lambert, Pyatkin, Morrison, Miranda, Lin, Chandu,
  Dziri, Kumar, Zick, Choi, Smith, and Hajishirzi}]{RewardBench}
Nathan Lambert, Valentina Pyatkin, Jacob Morrison, LJ~Miranda, Bill~Yuchen Lin,
  Khyathi Chandu, Nouha Dziri, Sachin Kumar, Tom Zick, Yejin Choi, Noah~A.
  Smith, and Hannaneh Hajishirzi. 2024.
\newblock \href {https://huggingface.co/spaces/allenai/reward-bench}
  {Rewardbench: Evaluating reward models for language modeling}.

\bibitem[{Li et~al.(2024{\natexlab{a}})Li, Dong, Tang, Wang, Zhang, Huang,
  Huang, Huang, Huang, Zhang et~al.}]{li2024synthetic}
Haoran Li, Qingxiu Dong, Zhengyang Tang, Chaojun Wang, Xingxing Zhang, Haoyang
  Huang, Shaohan Huang, Xiaolong Huang, Zeqiang Huang, Dongdong Zhang, et~al.
  2024{\natexlab{a}}.
\newblock Synthetic data (almost) from scratch: Generalized instruction tuning
  for language models.
\newblock \emph{arXiv preprint arXiv:2402.13064}.

\bibitem[{Li et~al.(2024{\natexlab{b}})Li, Chen, Chen, He, Gu, and
  Zhou}]{li2024selective}
Ming Li, Lichang Chen, Jiuhai Chen, Shwai He, Jiuxiang Gu, and Tianyi Zhou.
  2024{\natexlab{b}}.
\newblock Selective reflection-tuning: {S}tudent-selected data recycling for
  llm instruction-tuning.
\newblock \emph{arXiv preprint arXiv:2402.10110}.

\bibitem[{Li et~al.(2024{\natexlab{c}})Li, Zhang, He, Li, Zhao, Wang, Cheng,
  and Zhou}]{li2024superfiltering}
Ming Li, Yong Zhang, Shwai He, Zhitao Li, Hongyu Zhao, Jianzong Wang, Ning
  Cheng, and Tianyi Zhou. 2024{\natexlab{c}}.
\newblock \href {https://doi.org/10.18653/v1/2024.acl-long.769}
  {Superfiltering: Weak-to-strong data filtering for fast instruction-tuning}.
\newblock In \emph{Proceedings of the 62nd Annual Meeting of the Association
  for Computational Linguistics (Volume 1: Long Papers)}, pages 14255--14273,
  Bangkok, Thailand. Association for Computational Linguistics.

\bibitem[{Li et~al.(2024{\natexlab{d}})Li, Zhang, Li, Chen, Chen, Cheng, Wang,
  Zhou, and Xiao}]{li2023quantity}
Ming Li, Yong Zhang, Zhitao Li, Jiuhai Chen, Lichang Chen, Ning Cheng, Jianzong
  Wang, Tianyi Zhou, and Jing Xiao. 2024{\natexlab{d}}.
\newblock \href {https://doi.org/10.18653/v1/2024.naacl-long.421} {From
  quantity to quality: Boosting {LLM} performance with self-guided data
  selection for instruction tuning}.
\newblock In \emph{Proceedings of the 2024 Conference of the North American
  Chapter of the Association for Computational Linguistics: Human Language
  Technologies (Volume 1: Long Papers)}, pages 7602--7635, Mexico City, Mexico.
  Association for Computational Linguistics.

\bibitem[{Li et~al.(2024{\natexlab{e}})Li, Chiang, Frick, Dunlap, Zhu,
  Gonzalez, and Stoica}]{arenahard2024}
Tianle Li, Wei-Lin Chiang, Evan Frick, Lisa Dunlap, Banghua Zhu, Joseph~E.
  Gonzalez, and Ion Stoica. 2024{\natexlab{e}}.
\newblock \href {https://lmsys.org/blog/2024-04-19-arena-hard/} {From live data
  to high-quality benchmarks: The arena-hard pipeline}.

\bibitem[{Li et~al.(2023)Li, Zhang, Dubois, Taori, Gulrajani, Guestrin, Liang,
  and Hashimoto}]{alpaca_eval}
Xuechen Li, Tianyi Zhang, Yann Dubois, Rohan Taori, Ishaan Gulrajani, Carlos
  Guestrin, Percy Liang, and Tatsunori~B. Hashimoto. 2023.
\newblock Alpacaeval: An automatic evaluator of instruction-following models.
\newblock \url{https://github.com/tatsu-lab/alpaca_eval}.

\bibitem[{Liu and Zeng(2024)}]{skyworkreward2024}
Chris~Yuhao Liu and Liang Zeng. 2024.
\newblock \href {https://huggingface.co/Skywork} {Skywork reward model series}.
\newblock \url{https://huggingface.co/Skywork}.

\bibitem[{Liu et~al.(2024{\natexlab{a}})Liu, Liu, Wong, Li, Wang, Hu, and
  Zhang}]{liu2024selectit}
Liangxin Liu, Xuebo Liu, Derek~F Wong, Dongfang Li, Ziyi Wang, Baotian Hu, and
  Min Zhang. 2024{\natexlab{a}}.
\newblock Selectit: Selective instruction tuning for large language models via
  uncertainty-aware self-reflection.
\newblock \emph{arXiv preprint arXiv:2402.16705}.

\bibitem[{Liu et~al.(2024{\natexlab{b}})Liu, Wei, Liu, Si, Zhang, Rao, Zheng,
  Peng, Yang, Zhou et~al.}]{liu2024best}
Ruibo Liu, Jerry Wei, Fangyu Liu, Chenglei Si, Yanzhe Zhang, Jinmeng Rao,
  Steven Zheng, Daiyi Peng, Diyi Yang, Denny Zhou, et~al. 2024{\natexlab{b}}.
\newblock Best practices and lessons learned on synthetic data for language
  models.
\newblock \emph{arXiv preprint arXiv:2404.07503}.

\bibitem[{Liu et~al.(2023)Liu, Zeng, He, Jiang, and He}]{liu2023makes}
Wei Liu, Weihao Zeng, Keqing He, Yong Jiang, and Junxian He. 2023.
\newblock What makes good data for alignment? a comprehensive study of
  automatic data selection in instruction tuning.
\newblock \emph{arXiv preprint arXiv:2312.15685}.

\bibitem[{Luo et~al.(2023)Luo, Xu, Zhao, Sun, Geng, Hu, Tao, Ma, Lin, and
  Jiang}]{luo2023wizardcoder}
Ziyang Luo, Can Xu, Pu~Zhao, Qingfeng Sun, Xiubo Geng, Wenxiang Hu, Chongyang
  Tao, Jing Ma, Qingwei Lin, and Daxin Jiang. 2023.
\newblock Wizardcoder: Empowering code large language models with
  evol-instruct.
\newblock \emph{arXiv preprint arXiv:2306.08568}.

\bibitem[{Meta(2024{\natexlab{a}})}]{llama32}
Meta. 2024{\natexlab{a}}.
\newblock Llama-3.2-3b.
\newblock \url{https://huggingface.co/meta-llama/Llama-3.2-3B}.

\bibitem[{Meta(2024{\natexlab{b}})}]{llama32_webpage}
Meta. 2024{\natexlab{b}}.
\newblock \href
  {https://ai.meta.com/blog/llama-3-2-connect-2024-vision-edge-mobile-devices/}
  {Llama 3.2: Revolutionizing edge ai and vision with open, customizable
  models}.

\bibitem[{Meta(2024{\natexlab{c}})}]{llama31}
Meta. 2024{\natexlab{c}}.
\newblock Meet llama 3.1.
\newblock \url{https://llama.meta.com}.

\bibitem[{Muralidharan et~al.(2024)Muralidharan, Sreenivas, Joshi, Chochowski,
  Patwary, Shoeybi, Catanzaro, Kautz, and Molchanov}]{minitron2024}
Saurav Muralidharan, Sharath~Turuvekere Sreenivas, Raviraj Joshi, Marcin
  Chochowski, Mostofa Patwary, Mohammad Shoeybi, Bryan Catanzaro, Jan Kautz,
  and Pavlo Molchanov. 2024.
\newblock \href {https://arxiv.org/abs/2407.14679} {Compact language models via
  pruning and knowledge distillation}.
\newblock \emph{arXiv preprint arXiv:2407.14679}.

\bibitem[{OpenAI(2024)}]{tiktoken}
OpenAI. 2024.
\newblock Tiktoken.
\newblock \url{https://github.com/openai/tiktoken}.

\bibitem[{Sun et~al.(2023)Sun, Shen, Zhou, Zhang, Chen, Cox, Yang, and
  Gan}]{sun2024principle}
Zhiqing Sun, Yikang Shen, Qinhong Zhou, Hongxin Zhang, Zhenfang Chen, David
  Cox, Yiming Yang, and Chuang Gan. 2023.
\newblock Principle-driven self-alignment of language models from scratch with
  minimal human supervision.
\newblock \emph{Advances in Neural Information Processing Systems}, 36.

\bibitem[{Taori et~al.(2023)Taori, Gulrajani, Zhang, Dubois, Li, Guestrin,
  Liang, and Hashimoto}]{alpaca}
Rohan Taori, Ishaan Gulrajani, Tianyi Zhang, Yann Dubois, Xuechen Li, Carlos
  Guestrin, Percy Liang, and Tatsunori~B. Hashimoto. 2023.
\newblock Stanford alpaca: An instruction-following llama model.
\newblock \url{https://github.com/tatsu-lab/stanford_alpaca}.

\bibitem[{Team et~al.(2024)Team, Riviere, Pathak, Sessa, Hardin, Bhupatiraju,
  Hussenot, Mesnard, Shahriari, Ram{\'e} et~al.}]{team2024gemma}
Gemma Team, Morgane Riviere, Shreya Pathak, Pier~Giuseppe Sessa, Cassidy
  Hardin, Surya Bhupatiraju, L{\'e}onard Hussenot, Thomas Mesnard, Bobak
  Shahriari, Alexandre Ram{\'e}, et~al. 2024.
\newblock Gemma 2: Improving open language models at a practical size.
\newblock \emph{arXiv preprint arXiv:2408.00118}.

\bibitem[{Team(2024)}]{qwen2.5}
Qwen Team. 2024.
\newblock \href {https://qwenlm.github.io/blog/qwen2.5/} {Qwen2.5: A party of
  foundation models}.

\bibitem[{Teknium(2023)}]{OpenHermes2.5}
Teknium. 2023.
\newblock \href {https://huggingface.co/datasets/teknium/OpenHermes-2.5}
  {Openhermes 2.5: An open dataset of synthetic data for generalist llm
  assistants}.

\bibitem[{Touvron et~al.(2023)Touvron, Martin, Stone, Albert, Almahairi,
  Babaei, Bashlykov, Batra, Bhargava, Bhosale et~al.}]{touvron2023llama}
Hugo Touvron, Louis Martin, Kevin Stone, Peter Albert, Amjad Almahairi, Yasmine
  Babaei, Nikolay Bashlykov, Soumya Batra, Prajjwal Bhargava, Shruti Bhosale,
  et~al. 2023.
\newblock Llama 2: Open foundation and fine-tuned chat models.
\newblock \emph{arXiv preprint arXiv:2307.09288}.

\bibitem[{Wang et~al.(2024{\natexlab{a}})Wang, Xiong, Xie, Zhao, and
  Zhang}]{ArmoRM}
Haoxiang Wang, Wei Xiong, Tengyang Xie, Han Zhao, and Tong Zhang.
  2024{\natexlab{a}}.
\newblock Interpretable preferences via multi-objective reward modeling and
  mixture-of-experts.
\newblock In \emph{EMNLP}.

\bibitem[{Wang et~al.(2023)Wang, Kordi, Mishra, Liu, Smith, Khashabi, and
  Hajishirzi}]{wang-etal-2023-self-instruct}
Yizhong Wang, Yeganeh Kordi, Swaroop Mishra, Alisa Liu, Noah~A. Smith, Daniel
  Khashabi, and Hannaneh Hajishirzi. 2023.
\newblock Self-instruct: Aligning language models with self-generated
  instructions.
\newblock In \emph{Proceedings of the 61st Annual Meeting of the Association
  for Computational Linguistics (Volume 1: Long Papers)}, pages 13484--13508,
  Toronto, Canada. Association for Computational Linguistics.

\bibitem[{Wang et~al.(2024{\natexlab{b}})Wang, Li, Perot, Le, Miao, Zhang, Lee,
  and Pfister}]{wang2024codeclm}
Zifeng Wang, Chun-Liang Li, Vincent Perot, Long~T Le, Jin Miao, Zizhao Zhang,
  Chen-Yu Lee, and Tomas Pfister. 2024{\natexlab{b}}.
\newblock Codeclm: Aligning language models with tailored synthetic data.
\newblock \emph{arXiv preprint arXiv:2404.05875}.

\bibitem[{Xia et~al.(2024{\natexlab{a}})Xia, Malladi, Gururangan, Arora, and
  Chen}]{xia2024less}
Mengzhou Xia, Sadhika Malladi, Suchin Gururangan, Sanjeev Arora, and Danqi
  Chen. 2024{\natexlab{a}}.
\newblock Less: Selecting influential data for targeted instruction tuning.
\newblock \emph{arXiv preprint arXiv:2402.04333}.

\bibitem[{Xia et~al.(2024{\natexlab{b}})Xia, Yu, Dang, Yang, Wu, Tian, Chang,
  and Lin}]{xia2024rethinkingdataselectionscale}
Tingyu Xia, Bowen Yu, Kai Dang, An~Yang, Yuan Wu, Yuan Tian, Yi~Chang, and
  Junyang Lin. 2024{\natexlab{b}}.
\newblock \href {https://arxiv.org/abs/2410.09335} {Rethinking data selection
  at scale: Random selection is almost all you need}.
\newblock \emph{Preprint}, arXiv:2410.09335.

\bibitem[{Xu et~al.(2023{\natexlab{a}})Xu, Sun, Zheng, Geng, Zhao, Feng, Tao,
  and Jiang}]{xu2023wizardlm}
Can Xu, Qingfeng Sun, Kai Zheng, Xiubo Geng, Pu~Zhao, Jiazhan Feng, Chongyang
  Tao, and Daxin Jiang. 2023{\natexlab{a}}.
\newblock Wizardlm: Empowering large language models to follow complex
  instructions.
\newblock \emph{arXiv preprint arXiv:2304.12244}.

\bibitem[{Xu et~al.(2023{\natexlab{b}})Xu, Guo, Duan, and
  McAuley}]{xu-etal-2023-baize}
Canwen Xu, Daya Guo, Nan Duan, and Julian McAuley. 2023{\natexlab{b}}.
\newblock \href {https://doi.org/10.18653/v1/2023.emnlp-main.385} {Baize: An
  open-source chat model with parameter-efficient tuning on self-chat data}.
\newblock In \emph{Proceedings of the 2023 Conference on Empirical Methods in
  Natural Language Processing}, pages 6268--6278, Singapore. Association for
  Computational Linguistics.

\bibitem[{Xu et~al.(2024)Xu, Jiang, Niu, Deng, Poovendran, Choi, and
  Lin}]{xu2024magpie}
Zhangchen Xu, Fengqing Jiang, Luyao Niu, Yuntian Deng, Radha Poovendran, Yejin
  Choi, and Bill~Yuchen Lin. 2024.
\newblock Magpie: Alignment data synthesis from scratch by prompting aligned
  llms with nothing.
\newblock \emph{arXiv preprint arXiv:2406.08464}.

\bibitem[{Yang et~al.(2024)Yang, Yang, Hui, Zheng, Yu, Zhou, Li, Li, Liu,
  Huang, Dong, Wei, Lin, Tang, Wang, Yang, Tu, Zhang, Ma, Xu, Zhou, Bai, He,
  Lin, Dang, Lu, Chen, Yang, Li, Xue, Ni, Zhang, Wang, Peng, Men, Gao, Lin,
  Wang, Bai, Tan, Zhu, Li, Liu, Ge, Deng, Zhou, Ren, Zhang, Wei, Ren, Fan, Yao,
  Zhang, Wan, Chu, Liu, Cui, Zhang, and Fan}]{qwen2}
An~Yang, Baosong Yang, Binyuan Hui, Bo~Zheng, Bowen Yu, Chang Zhou, Chengpeng
  Li, Chengyuan Li, Dayiheng Liu, Fei Huang, Guanting Dong, Haoran Wei, Huan
  Lin, Jialong Tang, Jialin Wang, Jian Yang, Jianhong Tu, Jianwei Zhang,
  Jianxin Ma, Jin Xu, Jingren Zhou, Jinze Bai, Jinzheng He, Junyang Lin, Kai
  Dang, Keming Lu, Keqin Chen, Kexin Yang, Mei Li, Mingfeng Xue, Na~Ni, Pei
  Zhang, Peng Wang, Ru~Peng, Rui Men, Ruize Gao, Runji Lin, Shijie Wang, Shuai
  Bai, Sinan Tan, Tianhang Zhu, Tianhao Li, Tianyu Liu, Wenbin Ge, Xiaodong
  Deng, Xiaohuan Zhou, Xingzhang Ren, Xinyu Zhang, Xipin Wei, Xuancheng Ren,
  Yang Fan, Yang Yao, Yichang Zhang, Yu~Wan, Yunfei Chu, Yuqiong Liu, Zeyu Cui,
  Zhenru Zhang, and Zhihao Fan. 2024.
\newblock Qwen2 technical report.
\newblock \emph{arXiv preprint arXiv:2407.10671}.

\bibitem[{Zar(2005)}]{zar2005spearman}
Jerrold~H Zar. 2005.
\newblock Spearman rank correlation.
\newblock \emph{Encyclopedia of Biostatistics}, 7.

\bibitem[{Zhang et~al.(2023)Zhang, Dong, Li, Zhang, Sun, Wang, Li, Hu, Zhang,
  Wu et~al.}]{zhang2023instruction}
Shengyu Zhang, Linfeng Dong, Xiaoya Li, Sen Zhang, Xiaofei Sun, Shuhe Wang,
  Jiwei Li, Runyi Hu, Tianwei Zhang, Fei Wu, et~al. 2023.
\newblock Instruction tuning for large language models: A survey.
\newblock \emph{arXiv preprint arXiv:2308.10792}.

\bibitem[{Zhao et~al.(2024)Zhao, Ren, Hessel, Cardie, Choi, and
  Deng}]{zhao2024wildchat}
Wenting Zhao, Xiang Ren, Jack Hessel, Claire Cardie, Yejin Choi, and Yuntian
  Deng. 2024.
\newblock \href {https://openreview.net/forum?id=Bl8u7ZRlbM} {Wildchat: 1m
  chat{GPT} interaction logs in the wild}.
\newblock In \emph{The Twelfth International Conference on Learning
  Representations}.

\bibitem[{Zheng et~al.(2024)Zheng, Chiang, Sheng, Li, Zhuang, Wu, Zhuang, Li,
  Lin, Xing, Gonzalez, Stoica, and Zhang}]{zheng2024lmsyschatm}
Lianmin Zheng, Wei-Lin Chiang, Ying Sheng, Tianle Li, Siyuan Zhuang, Zhanghao
  Wu, Yonghao Zhuang, Zhuohan Li, Zi~Lin, Eric Xing, Joseph~E. Gonzalez, Ion
  Stoica, and Hao Zhang. 2024.
\newblock \href {https://openreview.net/forum?id=BOfDKxfwt0} {{LMSYS}-chat-1m:
  A large-scale real-world {LLM} conversation dataset}.
\newblock In \emph{The Twelfth International Conference on Learning
  Representations}.

\end{thebibliography}

\clearpage
\appendix
\section{More on Experimental Setups}

\subsection{Instruction Set Details}
\label{appendix: instruction data details}

\begin{figure}[!h]
    \centering
    \includegraphics[width=0.9\linewidth]{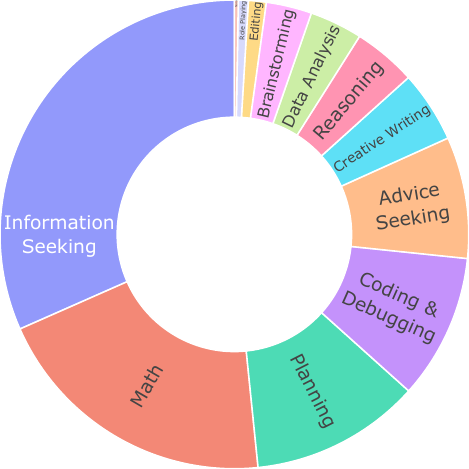}
    \caption{Task categories of the Magpie-100K instruction set used in our study.}
    \label{fig:instruction-tags}
\end{figure}

Figure \ref{fig:instruction-tags} demonstrates the task category of instructions in our sampled Magpie-100K. We follow \citep{xu2024magpie} and use \emph{Llama-3-8B-Instruct} to tag the task categories.
We note that this instruction set covers wide range of instructions across different task categories.

\subsection{Supervised Fine-Tuning Setups}
\label{appendix: Instruction-Tuning Setups}

Table \ref{tab: fine-tune hyperparameters} demonstrates the detailed supervised fine-tuning (SFT) hyper-parameters. We perform experiments on a server with four NVIDIA A100-SXM4-80GB GPUs, an AMD EPYC 7763 64-Core Processor, and 512 GB of RAM. These experiments were conducted using Axolotl\footnote{\url{https://github.com/OpenAccess-AI-Collective/axolotl}}.

\begin{table}[htbp]
\small
\centering
\caption{This table shows the hyper-parameters for supervised fine-tuning.}
\vspace{1em}
\begin{tabular}{ll}
\toprule
\textbf{Hyper-parameter} & \textbf{Value} \\ \midrule
Learning Rate & $2 \times 10^{-5}$ \\
Number of Epochs & $2$ \\
Number of Devices & $4$ \\
Per-device Batch Size & $1$ \\
Gradient Accumulation Steps & $8$ \\
Effective Batch Size & $32$ \\
Optimizer & \texttt{Adamw} \\
Learning Rate Scheduler & \texttt{cosine} \\
Warmup Steps & $100$ \\
Max Sequence Length  & $4096$ \\ \bottomrule
\end{tabular}
\label{tab: fine-tune hyperparameters}
\end{table}

\section{More Experimental Results}

\subsection{Detailed Benchmark Scores of Instruction-Tuned LLMs}
\label{appendix: detailed benchmark scores}

Table \ref{table: full results} details the benchmark scores of AE2 and AH when tuning base models with different response generators. These results complement the Average Performance shown in Figure \ref{fig:main-base-response-eval}. 

\subsection{Larger Models' Paradox in Larger Base Models}
\label{appendix: larger base model}

We summarize the benchmark scores of AE2 and AH when tuning large base model (Llama-3.1-8B) with diverse
response generators in Table \ref{tab: large base models}. We observe that the Larger Models' Paradox persists when employing the Qwen2.5 and Llama-3.1 model families as response generators.
We further demonstrate that the Larger Model's Paradox is not an effect of data randomness in Table \ref{tab: data randomness}.

\begin{table}[htbp]
    \centering
    \caption{This table presents benchmark scores of AE2 and AH when tuning large base model (Llama-3.1-8B) with diverse
response generators. The Larger Models' Paradox persists when employing the Qwen2.5 and Llama-3.1 model families as response generators.}
\resizebox{0.5\textwidth}{!}{
    \begin{tabular}{l l c c c c}
        \toprule
        Base Model & Response Generator & AE2 LC & AE2 WR & AH & AP \\
        \midrule
        \multirow{11}{*}{Llama-3.1-8B} 
        & Qwen2.5-3B-Instruct & 11.48 & 13.85 & 15.90 & 13.74 \\
        & Qwen2.5-7B-Instruct & 18.70 & 20.22 & 25.90 & 21.61 \\
        & Qwen2.5-14B-Instruct & 17.50 & 17.19 & 28.60 & 21.10 \\
        & Qwen2.5-32B-Instruct & 16.20 & 16.42 & 27.80 & 20.14 \\
        & Qwen2.5-72B-Instruct & 29.73 & 32.35 & 30.90 & 30.99 \\
        & Llama-3.1-8B-Instruct & 12.62 & 14.34 & 15.80 & 14.25 \\
        & Llama-3.1-70B-Instruct & 14.98 & 17.74 & 21.00 & 17.91 \\
        & Llama-3.1-405B-Instruct & 15.40 & 17.00 & 16.50 & 16.30 \\
        & Gemma-2-2b-it & 17.11 & 19.64 & 15.60 & 17.45 \\
        & Gemma-2-9b-it & 25.74 & 22.88 & 23.40 & 24.00 \\
        & Gemma-2-27b-it & 25.09 & 24.60 & 25.40 & 25.00 \\
        \bottomrule
    \end{tabular}
}
\vspace{-1em}
    \label{tab: large base models}
\end{table}

\subsection{Impact of Data Randomness on Evaluation}
\label{appendix: data randomness}

We sample 80K instructions from Magpie-100K using different seeds and fine-tuned Llama-3.1-Minitron-4B. 
The performance of fine-tuned models is shown in Table \ref{tab: data randomness}.
We observe that the average performance varies by only 2.89\%, demonstrating that our evaluation is robust across different instruction samples. This finding underscores the consistency of our evaluation.

\begin{table*}[htbp]
\small
	\centering
	\caption{This table details benchmark scores of AE2 and AH when tuning different base models with diverse response generators.} 
\resizebox{\textwidth}{!}{
\begin{tabular}{c r rrrrrrrrrrrrrrrrrrrrrrr}
    \toprule
 \multirow{2}[2]{*}{Base Model} & \multirow{2}[2]{*}{Metric}  & \multicolumn{3}{c}{Phi-3} & \multicolumn{3}{c}{Gemma 2}  &\multicolumn{2}{c}{Llama 3} & \multicolumn{3}{c}{Llama 3.1} & \multicolumn{3}{c}{Qwen2} & \multicolumn{5}{c}{Qwen2.5}\\\cmidrule(lr){3-5} \cmidrule(lr){6-8} \cmidrule(lr){9-10} \cmidrule(lr){11-13} \cmidrule(lr){14-16} \cmidrule(lr){17-21}
 & & Mini & Small & Medium & 2B & 9B & 27B &  8B & 70B &  8B & 70B & 405B &  1.5B & 7B & 72B & 3B & 7B & 14B & 32B & 72B \\ 
\midrule
\multirow{3}[2]{*}{Qwen2-1.5B} & AE 2 WR & 3.65 & 3.64 & 2.80 & 5.34 & 6.13 & 5.49 & 3.39 & 3.74 & 2.76 & 3.49 & 3.09 & 2.83 & 4.09 & 3.35 & 5.60 & 6.84 & 5.13 & 5.65 & 7.03  \\
 & AE 2 LC & 2.85 & 2.98 & 2.18 & 4.16 & 5.60 & 4.99 & 2.64 & 3.10 & 2.10 & 2.74 & 2.36 & 2.68 & 3.47 & 2.82 & 4.50 & 5.66 & 4.38 & 4.96 & 5.83\\ \cmidrule{2-21}
 & AH & 1.8 & 1.8 & 1.2 & 4.4 & 5.2 & 4.5 & 1.9 & 2.6 & 2.2 & 2.8 & 2.4 & 1.0 & 3.3 & 1.8 & 2.6 & 4.3 & 4.4 & 3.7 & 4.8\\
 \midrule
 \multirow{3}[2]{*}{Gemma 2-2B} & AE 2 WR & 6.60 & 6.54 & 4.54 & 16.88 & 11.83 & 12.09 & 7.09 & 8.49 & 7.20 & 9.45 & 8.92 & 2.14 & 7.11 & 6.07 & 7.91 & 12.00 & 8.07 & 9.19 & 16.68\\
 & AE 2 LC & 5.90 & 5.89 & 3.99 & 12.93 & 12.51 & 13.09 & 5.70 & 7.13 & 5.63 & 7.32 & 7.11 & 1.91 & 6.45 & 5.46 & 6.84 & 10.94 & 7.53 & 8.77 & 13.85\\ \cmidrule{2-21}
 & AH & 3.3 & 4.1 & 2.6 & 12.9 & 9.3 & 9.9 & 5.2 & 5.6 & 4.9 & 5.8 & 5.8 & 0.9 & 5.7 & 3.4 & 6.5 & 7.1 & 8.4 & 6.9 & 9.6\\
\midrule
\multirow{3}[2]{*}{Qwen2.5-3B} & AE 2 WR & 8.19 & 7.79 & 5.97 & 10.52 & 13.57 & 10.01 & 8.07 & 10.17 & 7.91 & 9.68 & 9.12 & 2.98 & 8.54 & 6.86 & 16.22 & 12.76 & 10.32 & 11.71 & 18.42\\
 & AE 2 LC & 7.22 & 7.29 & 5.49 & 9.58 & 13.78 & 10.18 & 7.85 & 9.37 & 7.22 & 8.94 & 8.59 & 2.54 & 7.98 & 6.59 & 14.79 & 11.89 & 10.28 & 11.65 & 16.41\\ \cmidrule{2-21}
& AH & 10.5 & 11.0 & 8.3 & 11.8 & 19.4 & 19.6 & 9.7 & 11.4 & 10.9 & 13.8 & 12.7 & 2.1 & 14.4 & 10.6 & 24.8 & 20.4 & 17.9 & 19.9 & 21.2 \\ \midrule
\multirow{3}[2]{*}{\makecell{Llama-3.2-3B}} & AE 2 WR & 4.88 & 3.54 & 3.05 & 8.89 & 11.45 & 10.58 & 4.67 & 5.45 & 4.26 & 6.68 & 6.44 & 1.72 & 6.23 & 5.13 & 6.09 & 7.72 & 6.82 & 7.10 & 12.12\\
 & AE 2 LC & 4.11 & 2.95 & 2.37 & 7.49 & 10.60 & 9.79 & 3.79 & 4.52 & 3.17 & 5.19 & 5.17 & 1.28 & 5.41 & 4.49 & 5.11 & 6.63 & 5.92 & 6.32 & 9.99\\ \cmidrule{2-21}
 & AH &3.3 & 4.1 & 2.6 & 9.0 & 10.9 & 8.5 & 5.1 & 6.5 & 3.6 & 5.7 & 5.3 & 0.6 & 5.6 & 4.0 & 7.2 & 9.8 & 9.5 & 8.9 & 10.8\\ \midrule
\multirow{3}[2]{*}{\makecell{Llama-3.1-\\Minitron-4B}} & AE 2 WR & 6.35 & 7.11 & 4.83 & 11.80 & 14.50 & 11.90 & 6.11 & 9.87 & 8.24 & 9.61 & 10.03 & 2.30 & 7.84 & 8.45 & 10.27 & 12.05 & 11.30 & 11.65 & 19.58\\
 & AE 2 LC & 5.74 & 6.61 & 4.31 & 10.37 & 16.13 & 12.34 & 4.80 & 8.93 & 6.96 & 8.52 & 9.23 & 2.03 & 7.31 & 8.11 & 9.17 & 11.12 & 10.89 & 11.13 & 17.77\\ \cmidrule{2-21}
 & AH &3.9 & 4.5 & 3.6 & 10.7 & 11.0 & 11.9 & 4.7 & 6.0 & 6.0 & 5.6 & 6.2 & 0.9 & 6.4 & 5.1 & 8.3 & 9.2 & 11.1 & 10.2 & 12.2\\
\bottomrule
\end{tabular}
}
\label{table: full results}
\end{table*}


\subsection{Visualization of baseline methods in measuring the effectiveness of response generators}
\label{appendix: more baseline visualization}

\begin{figure}[htbp]
    \centering
    \includegraphics[width=0.5\textwidth]{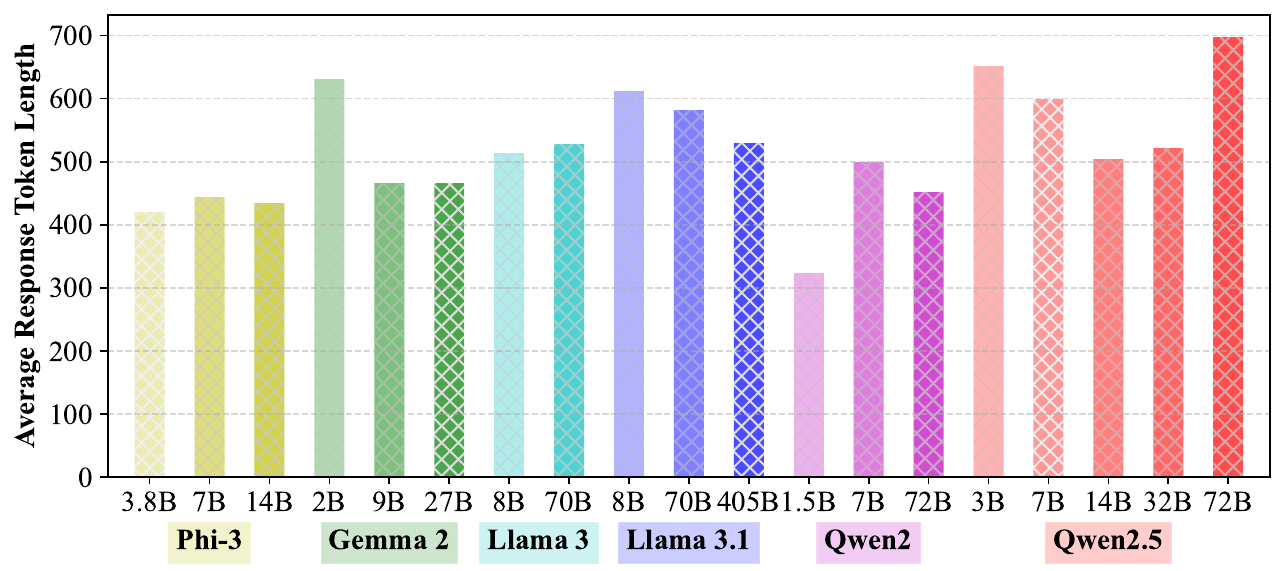}
    \caption{Average Output Length of synthetic datasets generated using different response generators (measured in Tokens).}
    \label{fig:model-Length}
    \vspace{-0.5em}
\end{figure}

\begin{figure}[htbp]
    \centering
    \includegraphics[width=0.5\textwidth]{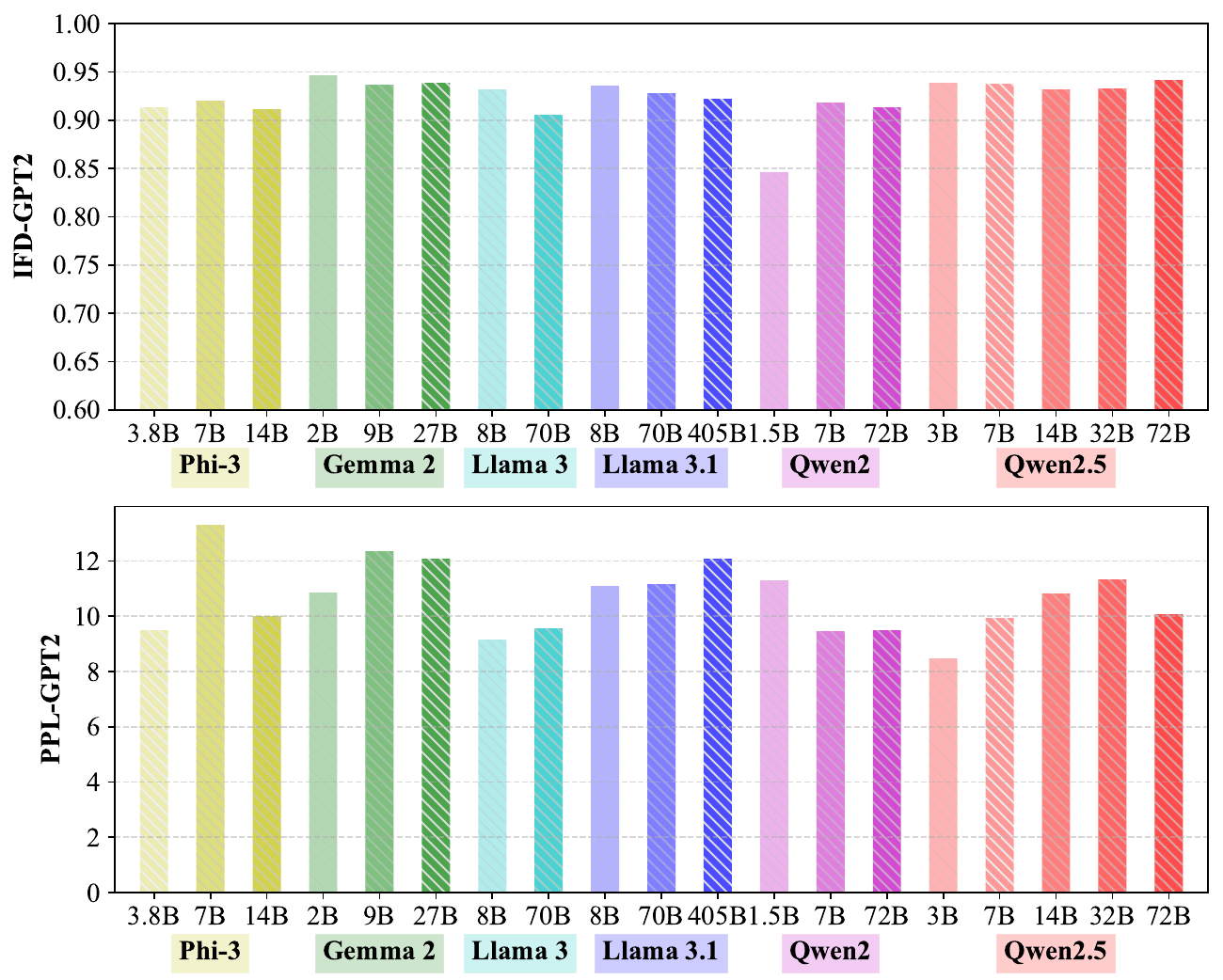}
    \caption{PPL-GPT2 and IFD-GPT2 of synthetic datasets generated using different response generators.}
    \label{fig:model-GPT2}
    \vspace{-0.5em}
\end{figure}

\begin{figure}[!t]
    \centering
    \includegraphics[width=0.5\textwidth]{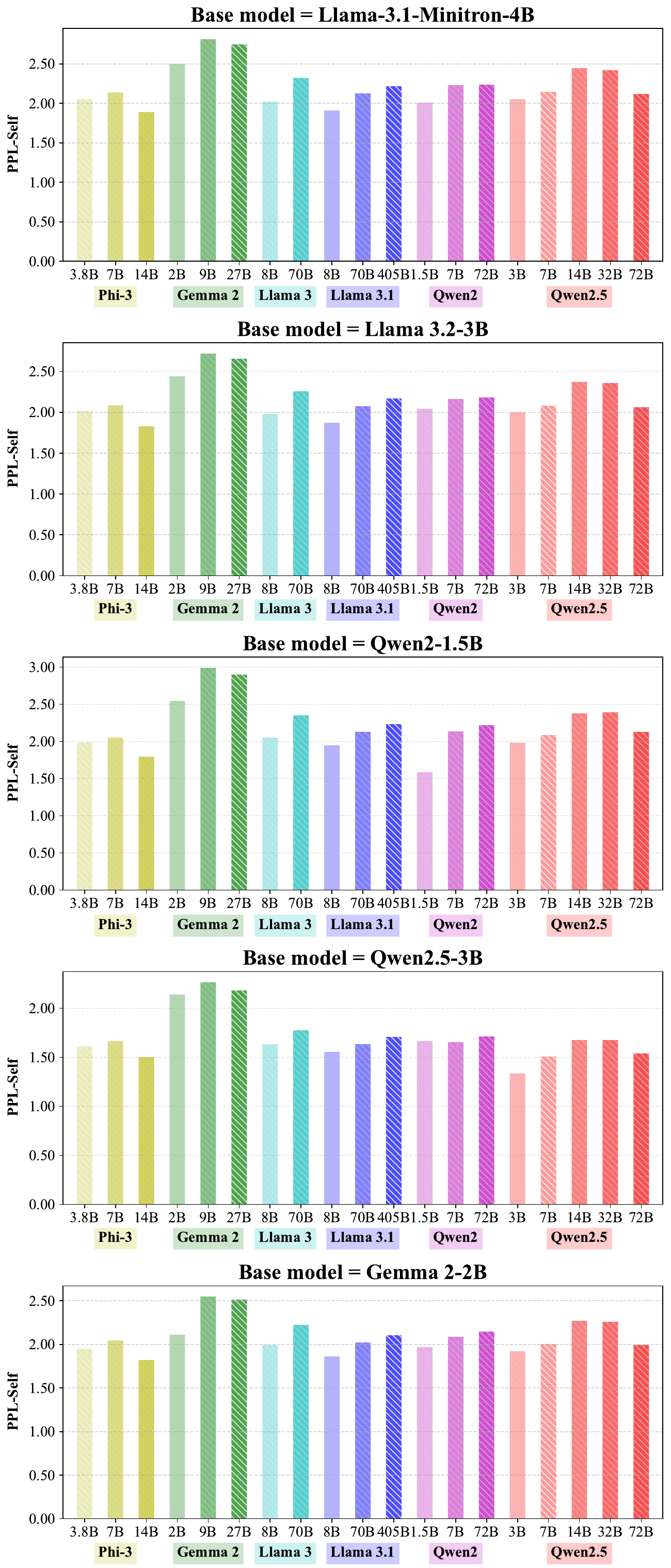}
    \caption{PPL-Self of five base models.}
    \label{fig:model-PPL}
\end{figure}

\begin{figure}[!t]
    \centering
    \includegraphics[width=0.5\textwidth]{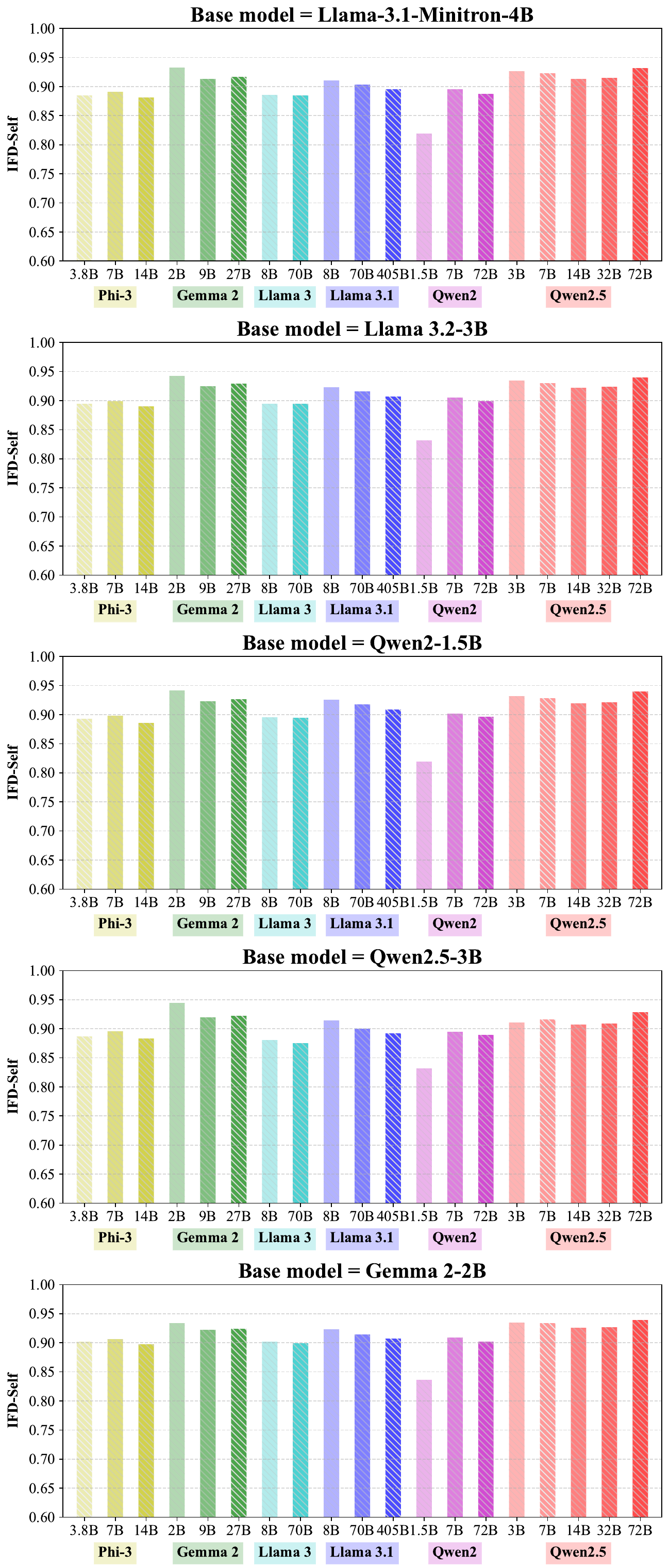}
    \caption{IFD-Self of five base models.}
    \label{fig:model-IFD}
\end{figure}

Figure \ref{fig:model-Length} presents the output length of synthetic datasets for each response generator. Figure \ref{fig:model-GPT2} visualizes the PPL-GPT2 and IFD-GPT2 across different response generators. Figure \ref{fig:model-PPL} and \ref{fig:model-IFD} reports PPL-Self and IFD-Self, respectively. We observe that although PPL-Self and IFD-Self have higher correlation compared with measuring using GPT2, they still to fail to effectively predict the effectiveness of different response generators, with low Spearman’s rank correlation coefficients demonstrated in Table \ref{tab: spearman}.

\begin{table}[htbp]
    \centering
    \caption{We sample 80K instructions from Magpie-100K using different seeds and fine-tuned Llama-3.1-Minitron-4B with the sampled data. We observe that the average performance varies by only 2.89\%, demonstrating that our evaluation is robust across different instruction samples. This finding underscores the consistency of our evaluation.}
    \resizebox{0.5\textwidth}{!}{
    \begin{tabular}{l c c c c}
        \toprule
        Instruction Sample & AE2 LC & AE2 WR & AH & Average Performance \\
        \midrule
        Magpie-80K (Seed = 42)  & 14.26  & 13.54  & 12.50  & 13.433 \\
        Magpie-80K (Seed = 123) & 13.40  & 12.92  & 12.80  & 13.040 \\
        Magpie-80K (Seed = 456) & 14.77  & 12.98  & 11.10  & 12.950 \\
        Magpie-80K (Seed = 789) & 13.57  & 12.79  & 11.20  & 12.520 \\
        \midrule
        Average & 14.00  & 13.058  & 11.90  & 12.986 \\
        Standard Deviation & 0.634  & 0.331  & 0.876  & 0.375 \\
        \bottomrule
    \end{tabular}
    }
    \vspace{-0.5em}
    \label{tab: data randomness}
\end{table}

\subsection{Impact of Reward Models on the performance of CAR}

We perform ablation analysis on the choice of reward models with a weaker reward model, Skywork-Reward-Llama-3.1-8B, and calculate CAR. The Spearman's correlations are presented in Table \ref{tab: reward model ablation}.
We observe that CAR using the weaker Skywork 8B reward model performs worse compared to using the stronger Skywork 27B reward model, indicating the reliance of CAR on a good performing reward model. However, even with a weaker reward model, CAR outperforms compared with using the reward model alone.

\begin{table}[htbp]
\small
    \centering
    \caption{Spearman's correlations when CAR uses different reward models. CAR relies on a good reward model. However, even with a weaker reward model, CAR outperforms compared with using the reward model alone.}
    \resizebox{0.48\textwidth}{!}{
    \begin{tabular}{lcccc}
        \toprule
        \makecell{Base\\Model} & \makecell{Skywork\\8B} & \makecell{CAR\\(Skywork 8B)} & \makecell{Skywork\\27B} & \makecell{CAR\\(Skywork 27B)} \\
        \midrule
        Qwen2-1.5B & 0.7895 & 0.7474 & 0.8754 & \textbf{0.8842} \\
        Gemma 2-2B & 0.7982 & 0.8018 & 0.8842 & \textbf{0.9000} \\
        Qwen2.5-3B & 0.7351 & 0.7386 & 0.7456 & \textbf{0.8105} \\
        Llama-3.1-Minitron-4B & 0.8860 & 0.8912 & 0.9386 & \textbf{0.9439} \\
        Llama-3.2-3B & 0.8105 & 0.8105 & \textbf{0.9088} & 0.9053 \\
        \bottomrule 
    \end{tabular}
    }
    \vspace{-0.5em}
    \label{tab: reward model ablation}
\end{table}

\end{document}